\documentclass[journal,12pt,onecolumn,draftclsnofoot,]{IEEEtran}
\IEEEoverridecommandlockouts
\usepackage{cite}
\usepackage{amsmath,amssymb,amsfonts}
\usepackage{algorithmic}
\usepackage{graphicx}
\usepackage{textcomp}
\usepackage{xcolor}
\usepackage{subfig}
\usepackage{amsmath}
\usepackage[capitalise]{cleveref}
\usepackage{threeparttable}
\usepackage[T1]{fontenc}
\usepackage{tabularx}
\usepackage{enumitem}
\usepackage{booktabs}

\def\BibTeX{{\rm B\kern-.05em{\sc i\kern-.025em b}\kern-.08em
    T\kern-.1667em\lower.7ex\hbox{E}\kern-.125emX}}
\begin{document}

\title{Re-Identifying K\={a}k\={a} with AI-Automated Video Key Frame Extraction
}

\author{
    Paula~Maddigan,
    Andrew~Lensen, 
    Rachael C.\ Shaw
    \thanks{A.\ Lensen and R.\ Shaw were supported by the Endeavour Fund---Smart Ideas from the New Zealand Ministry of Business, Innovation \& Employment under contract VUW RTVU2442.}
	\thanks{P.\ Maddigan, A. Lensen are with the Centre for Data Science and Artificial Intelligence, and School of Engineering and Computer Science;  Victoria University of Wellington; Wellington 6140; New Zealand (e-mail: paula.maddigan@vuw.ac.nz; andrew.lensen@vuw.ac.nz) }%  
	\thanks{R.\ C.\ Shaw is with School of Biological Sciences;  Victoria University of Wellington; Wellington 6140; New Zealand (e-mail: rachael.shaw@vuw.ac.nz)  }%  

}

\maketitle

\begin{abstract}

 Accurate recognition and re-identification of individual animals is essential for successful wildlife population monitoring. Traditional methods, such as leg banding of birds, are time consuming and invasive. Recent progress in artificial intelligence, particularly computer vision, offers encouraging solutions for smart conservation and efficient automation. 
 This study presents a unique pipeline for extracting high-quality key frames from videos of k\={a}k\={a} (\textit{Nestor meridionalis}), a threatened forest-dwelling parrot in New Zealand. Key frame extraction is well-studied in person re-identification, however, its application to wildlife is limited.
Using video recordings %of wild k\={a}k\={a} 
at a custom-built feeder, we extract key frames and evaluate the re-identification performance of our pipeline. Our unsupervised methodology combines object detection using YOLO and Grounding DINO, optical flow blur detection, image encoding with DINOv2, and clustering methods to identify representative key frames.
The results indicate that our proposed key frame selection methods yield image collections which achieve high accuracy in k\={a}k\={a} re-identification, providing a foundation for future research using media collected in more diverse and challenging environments.
Through the use of artificial intelligence and computer vision, our non-invasive and efficient approach provides a valuable alternative to traditional physical tagging methods for recognising k\={a}k\={a} individuals and therefore improving the monitoring of populations. This research contributes to developing fresh approaches in wildlife monitoring, with applications in ecology and conservation biology.
\end{abstract}


\section{Introduction}
\IEEEPARstart{E}{ffective} conservation of wildlife populations requires understanding individual animal behaviour. How animals behave can influence population dynamics and a species role within an ecosystem  \cite{Anthony2000,Nathan2008}. Yet a detailed knowledge of behaviour is built on accurate recognition of individuals and their re-identification (re-ID) across locations and time. Animal re-ID therefore plays a vital part in conservation \cite{Stephenson2019,Debicki2021}.
However, many species present substantial re-ID challenges for researchers due to their complex social behaviours, high mobility, and visual similarities among individuals \cite{White2012, Vidal2021, Schneider2019}. For these reasons, the k\={a}k\={a} (\textit{Nestor meridionalis}), a sociable but endangered parrot found in select forests and urban areas in New Zealand, is an ideal example of a species that requires innovative monitoring approaches. Traditional methods for k\={a}k\={a} re-ID, such as leg banding and visual re-sightings of banded individuals, are time-consuming and invasive \cite{Li2020, Crouse2017}. Recent advancements in computer vision (CV) and machine learning (ML) have created new opportunities for automating the re-ID of individuals within a population, enabling researchers to collect high-quality data and video footage of individual animals without disrupting their natural behaviour \cite{Weinstein2018, Lurig2021, Borowiec2022, Miele2021,Tuia2022}.  

Selecting key frames (a compact set of representative images) from video is essential in preparing data for downstream analysis. It preserves information, reduces redundancy, improves computational efficiency, and optimises storage space. While this concept has been widely recognised and addressed across various CV research domains, its importance has been under-explored for recognising individuals in wildlife recordings despite, being equally relevant\cite{openanimals2024}. 
One reason for this lies in the nature of the available data. Research on person re-ID often leverages large datasets such as those collected from surveillance, security, and smart city solutions \cite{Gu2022,Selvan2025,karanam2019}, which provide an abundance of diverse data. Human movement tends to be predictable and key frames can be easily identified based on face position, gait, or scene changes.  
In contrast, for some wildlife species, their movements are unpredictable, and highly variable in pose and behaviour, unlike human subjects.  
 Moreover, wildlife re-ID is hindered by limited access to sizable data collections, relying instead on small species-specific datasets \cite{fintan2023,Maddigan2024,Xiao2023,Ferreira2020,Rogers2024,Zuerl2023,Schofield2023,Wang2021}. 
These limitations not only restrict the diversity across frames, but also underscores the need for effective frame selection strategies similar to those used in other re-ID tasks.  

This study proposes a methodology for automating key frame extraction from videos for wildlife re-ID, and we illustrate its application to k\={a}k\={a}. Using footage of k\={a}k\={a} at a custom-built feeder station, we develop an innovative 
pipeline that improves upon the accuracy of our previous heuristic-based approach \cite{fintan2023,Maddigan2024}. By leveraging recent developments in artificial intelligence (AI) models for object detection and image embedding, combined with traditional methods like clustering and optical flow, we present a robust unsupervised methodology for extracting feature-rich frames ideal for image matching.

Our work marks a significant contribution to research in k\={a}k\={a} monitoring and re-ID and offers transferable solutions that may be adapted for use with other wildlife species.
We hope our study will raise awareness about the importance of key frame extraction in wildlife re-ID pipelines for improving accuracy, mirroring its established significance for other downstream video analysis tasks such as classification and summarisation.

\subsection*{Contributions}
\begin{itemize}
    \item We propose an innovative approach for automating key frame extraction from videos to improve the accuracy in recognising individual k\={a}k\={a}.
    \item We develop and fine-tune an AI model for detecting k\={a}k\={a} in images.
    \item We create 
    a comprehensive dataset comprising of selected k\={a}k\={a} images which may serve as a foundation for future kākā re-ID studies and the development of new re-ID methods.
\end{itemize}

\section{Related Work}
\subsection{Key Frame Extraction from Videos}
Methods for extracting key frames from videos have been widely studied, especially for summarising and classifying videos
\cite{ Liu2022, SavranKiziltepe2023}. We briefly outline some 
recent approaches which have combined different methods to achieve promising results for these tasks before focusing on those specifically for wildlife re-ID. 
One study \cite{Kaur2024} filtered out unimportant frames by extracting only one frame per second and removing monochromatic frames. The authors then aggregated visual and structural features before clustering to select informative frames, concluding their pipeline performed well on OpenVideo and YouTube datasets.
In another approach \cite{Tang2023}, the authors used a CNN to extract features from the frames, which they subsequently clustered using a density peaks clustering algorithm to select key frames. They then used an LSTM for classifying the video, which showed promise on action recognition and human motion datasets.
Other researchers \cite{Tan2024} have used deep learning
to divide videos into shots and extract relevant features,
performed clustering, then remove redundant frames. The application of such hybrid approaches to key frame extraction from videos has successfully spanned various domains, from sports analysis \cite{Sarwar2023, Yan2022} to deepfake detection \cite{Mitra2021} and sign language recognition \cite{Athira2022}. 

Extracting key frames from videos is commonly used to represent the story within a video \cite{Tang2023}, documenting the pictorial narrative and scene changes. However, we aim in our study to extract key frames that provide high-clarity images for the downstream task of recognising individuals, capturing each bird's intricate and distinctive characteristics for accurate feature matching and re-ID.  
To date, significant research has focused on key frame extraction for person re-ID in videos \cite{Lu2022, Xie2022, Tao2023}. On the other hand, wildlife re-ID tasks have predominantly focused on image matching, where individual images are compared to measure similarity, rather than considering video clips as a whole \cite{cermak2024,Wahltinez2024, Adam_2024, Nepovinnykh2025, Matlala2025,Cermak2023, Ghaffari2024}. 

\subsection{Key Frame Extraction for Wildlife Re-ID}
Recently, a limited number of wildlife re-ID studies have explored video-based approaches, but these have neither considered,  nor automated the extraction of key frames. 
In a study on meerkats \cite{Rogers2024} the authors 
extracted one frame per second from their video clips and augmented the dataset with additional extractions by staggering the start interval. They identified the maximum dimensions of the animal's bounding box throughout a video clip and cropped all frames with those coordinates, generating images of a fixed-size square.

Researchers collected videos of polar bears in zoos \cite{Zuerl2023} and generated eight-second sequences at 12.5 fps, totalling 100 frames per clip. They used a YOLO object detection model to create bounding boxes and annotate the dataset. Applying a CNN deep learning model, they measured the re-ID accuracy using the extracted images and employed a video person re-ID model \cite{Li2019} for comparison. They concluded the video-based method outperformed the image comparison approach.

In a study of chimpanzees using video recordings captured in the wild, the authors extracted frames every 10 seconds. They manually annotated each image with bounding boxes and labels to identify the  23 individuals \cite{Schofield2023}.

Using streaming videos of Pandas \cite{Wang2021}, researchers calculated the similarity between adjacent frames using a custom metric, retaining only frames considered different from their predecessors. In addition, they manually added images with contrasting visual characteristics. Cropping of the pandas was done manually. 

Several studies have extracted frames from videos for the re-ID of birds \cite{fintan2023,Maddigan2024,Xiao2023,Ferreira2020}. In our previous work for the re-ID of k\=ak\=a \cite{fintan2023,Maddigan2024}, we used a heuristic approach and manually extracted key frames to collate our datasets based on videos. We first extracted every 10th frame. Then we ensured the bird's head and beak were fully visible inside the feeder within these frames and excluded any images we viewed as blurry or low-quality. We cropped the frames using static coordinates to remove unnecessary background details. 

A tracking and re-ID study on a flock of 15 cowbirds (\textit{Molothrus ater}) housed in an aviary \cite{Xiao2023} used a more dynamic approach to key frame extraction by virtually partitioning the aviary into 3D bins. The authors extracted every 10th frame from the video clips until the bin for each bird contained 10 images, upon which they subsequently reduced the extraction rate to every 40th frame. They note this method aided in ensuring a diversity of locations in their images. They utilised a mask R-CNN model for object detection and cropped the images based on the bounding boxes.

Another study on three small bird species \cite{Ferreira2020} chose to configure their cameras to capture an image at 2-second intervals instead of recording video. This approach prevented collecting nearly identical frames and eliminated the need for extracting key frames.

\subsection{Summary}
There is limited research on key frame extraction from videos for wildlife re-ID, particularly for species with subtle distinguishing features. Parrots, such as k\={a}k\={a}, pose a greater challenge than person re-ID due to the minimal visual differences in appearance among individuals. 
Furthermore, existing re-ID work often neglects the significance of frame selection, overlooking the importance of choosing sharp and clear images across diverse scenes and the potential for ML to detect subtle differences between frames that are imperceptible to humans.  
This gap highlights the need for specialised, automated AI-based approaches to key frame selection that can effectively capture the distinguishing traits of individuals within wildlife species.

\section{Study Data}
\begin{figure}[!h]
\centering
   \includegraphics[width=0.48\textwidth]{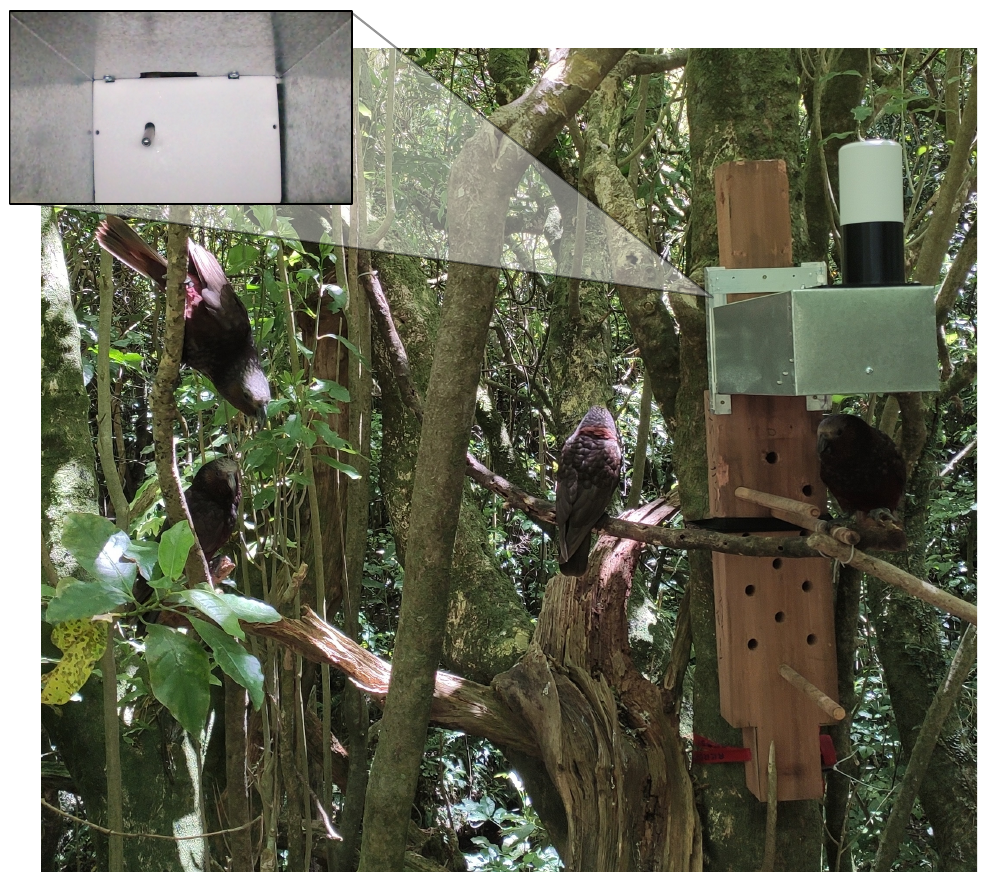}
    \caption{The custom-built k\={a}k\={a} feeder during an active period of feeding. The design features a ledge for the bird to perch on and a nozzle that dispenses food. The setup enables recording of the k\={a}k\={a}'s head in a profile view to optimise the capture of the unique beak morphology of individuals. The figure inset shows the white, non-reflective plastic cover installed inside the feeder to control the background environment.
  }
  \label{fig:feeder_station}
\end{figure}

We sourced our study data\footnote{The data collection process was approved by the Victoria University of Wellington Animal Ethics Committee (Approval Number 29656)} from video recordings captured at Zealandia Te Māra a Tāne, a predator-free urban ecosanctuary located in Wellington, New Zealand.
We installed a motion-detecting GoPro Hero camera inside a purpose-built feeding station at the ecosanctuary to record k\={a}k\={a} visits \cite{Maddigan2024,fintan2023}, as shown in \cref{fig:feeder_station}. 

Our dataset comprises three distinct collection periods: (A) November 2021, utilising a GoPro Hero 8 camera; (B) 15 November 2022 to 7 January 2023, with the same GoPro Hero 8 camera; and (C) 8 January 2023 to 18 January 2023, with an upgraded GoPro Hero 10. 

To enable us to evaluate our proposed re-ID pipeline, we observed and recorded the identity of  all k\={a}k\={a} that visited our feeding station during each data collection period. We identified known individuals via their leg-band combinations\footnote{A cohort colour band on one leg and up to two smaller coloured bands on the other. For example ``WS-P" indicates a white band above a silver band on the left leg and a pink one on the right. } \cite{Maddigan2024,fintan2023}. This approach generated an extensive dataset, predominantly comprising unlabelled (unbanded) birds, with a small proportion of labelled (banded) birds. We subsequently created three labelled datasets, one for each collection period, as described in detail in our earlier work \cite{Maddigan2024,fintan2023}. We developed our re-ID pipeline using an unsupervised framework to accommodate the identification of previously unseen individuals in future work. Therefore, to assess the accuracy of our system in identifying individual k\={a}k\={a}, we used these labelled datasets to establish accuracy metrics. \Cref{tab:ds_summary} lists the count of labelled birds in each dataset and the number of videos for these labelled birds.

\begin{table}[htbp]
%\vspace{-1em}
\caption{ Summary of Video Datasets}
\begin{center}
\begin{tabular}{crrrr}
\hline 
\noalign{\smallskip}
{Dataset} & {Labelled Birds}& 
{Videos}\vspace{2pt}\\
\hline \noalign{\smallskip}
A & 7 & 128	\\
B & 15 & 1,821	\\
C & 14&517 \\\noalign{\smallskip}
\hline
%\vspace{-2em}
\end{tabular}
\label{tab:ds_summary}
\end{center}
\end{table}

\section{Methodology}

\subsection{Overview}
\begin{figure}[!h]
\centerline{
%\vspace{-1em}
\includegraphics[width=0.98\textwidth]{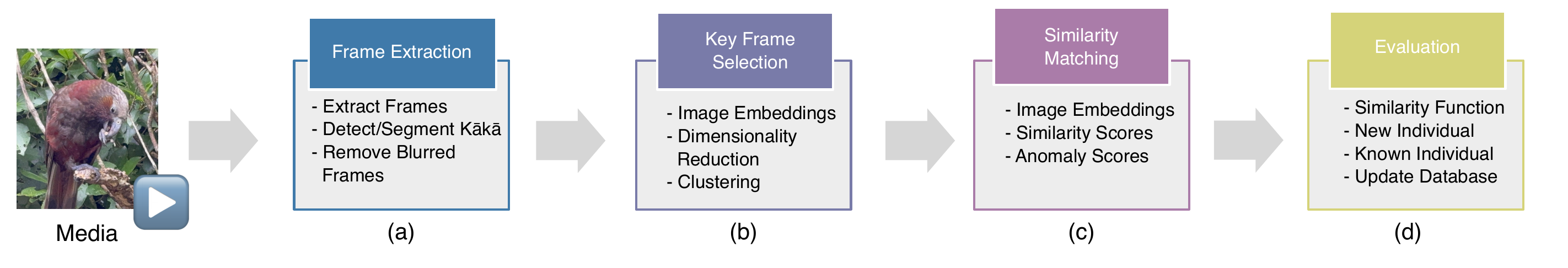}
}
\caption{Methodology for identifying an individual k\={a}k\={a} from visual media.}
\label{fig:pipleline_overview}
%\vspace{-1em}
\end{figure}

The workflow to identify a k\={a}k\={a} within visual media is illustrated in \cref{fig:pipleline_overview}. 
We designed the system as a pipeline of interconnected units to enable easy integration of new components or modify/replace existing ones.

\noindent The pipeline takes as input visual media such as video, still photos, or live photos and consists of four stages:

\begin{enumerate}[label=\textbf{(\alph*)}]
    \item \textbf{Frame Extraction}: Extract all frames from the visual media (for still photos there will be only one frame).  Create a candidate set of frame(s) by detecting and segmenting k\={a}k\={a} with a fine-tuned object detection/segmentation model, then retaining those frames with low blur.
    \item \textbf{Key Frame Selection}: Build embeddings to encode each candidate frame for media with more than a set number of minimum frames. Select key frames using methods such as clustering, with dimensionality reduction options during pre-processing.
    \item \textbf{Similarity Matching}: Build embeddings to encode each key frame. Calculate similarity scores and anomaly scores by comparing the key frame embeddings to database embeddings of known birds. Identify the labels of each best matched embedding.
    \item  \textbf{Evaluation}: Using the labels from the best matched embeddings, together with similarity scores and anomaly scores,  identify the k\={a}k\={a} in the visual media. Update the database with the new frame embeddings and labels.  
\end{enumerate}

In previous work using these video datasets \cite{Maddigan2024,fintan2023}, key frames were generated using an alternative heuristic-based method to that presented in this study. Hence, we have also performed the \textbf{Similarity Matching} and \textbf{Evaluation} stages using these heuristically generated key frame sets to provide a comparison.

In the next section we begin by introducing K\=ak\=a-YOLO, a fine-tuned YOLO model for k\=ak\=a object detection. This model is used during the \textbf{Frame Extraction} stage of the pipeline in \cref{fig:pipleline_overview}(a).
We then expand on each stage of the pipeline in subsequent sections.

\subsection{K\={a}k\={a}-YOLO: Fine-Tuned Object Detection Model}
\label{sec:kakayolo}
To improve the accuracy and efficiency of detecting k\={a}k\={a} in images during the frame extraction stage of the pipeline in \ref{fig:pipleline_overview}(a), we first fine-tuned a pre-trained YOLO (You Only Look Once) model. The data was captured with minimal background noise hence we chose to build the object detection component of the model without segmenting the bird. Initial experiments with segmentation did not show improved re-ID accuracy for these datasets. 
The YOLO algorithm is a state-of-the-art real-time object detection model well-suited for detecting objects in images and videos \cite{Redmon2016,khanam2024al}. It is based on the Convolutional Neural Network (CNN) architecture with layers for feature extraction and object detection. The model is pre-trained on the COCO (Common Objects in Context) dataset with approximately 300K images (200k with annotations) and 80 object categories such as cars, bicycles etc.  Images are passed through the deep neural network to predict bounding box coordinates to encapsulate objects and determine their class labels with probabilities.

We selected the YOLO11 model 
\cite{yolo11_ultralytics} which was the most recent version at the time of our study. We used the medium sized variant, YOLO11m. This model was recently measured against other variants and considered a top-performing model in accuracy, size and efficiency \cite{jegham2025}, offering an ideal solution for fast real-time detection. 

\begin{figure}[!h]
\centerline{
%\vspace{-1em}
\includegraphics[width=0.95\textwidth]{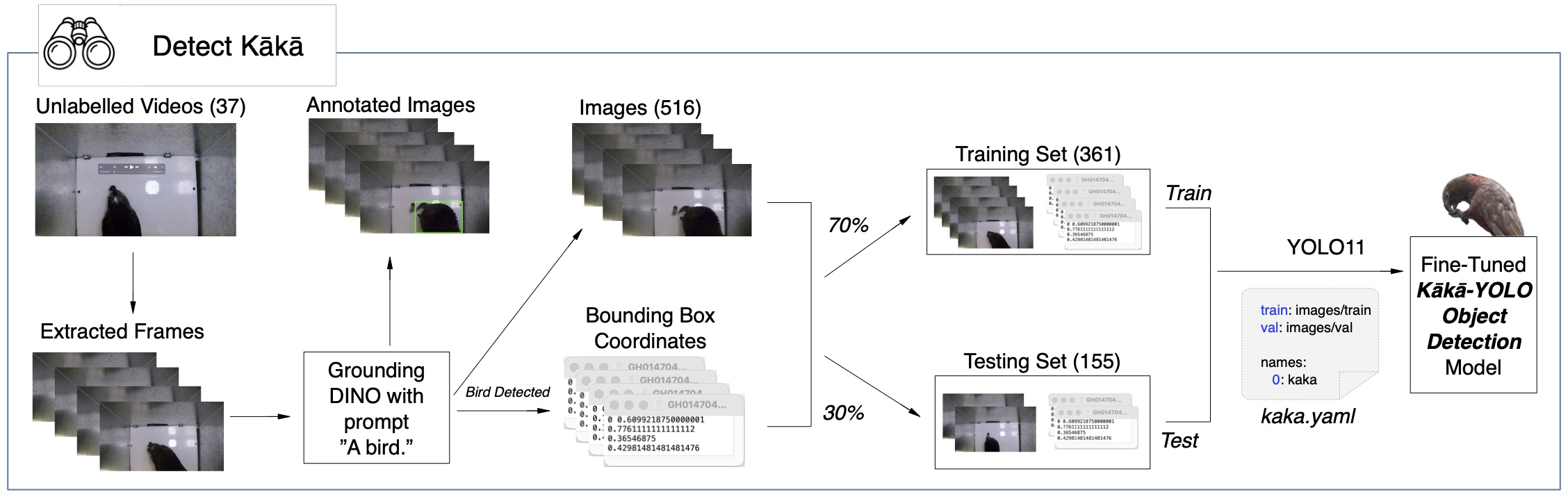}
}
\caption{Overview of the development of K\={a}k\={a}-YOLO, a fine-tuned YOLO model for k\={a}k\={a} object detection.}
\label{fig:yolo_model}
%\vspace{-1em}
\end{figure}

\begin{figure}[!h]
\centerline{
%\vspace{-1em}
\includegraphics[width=0.7\textwidth]{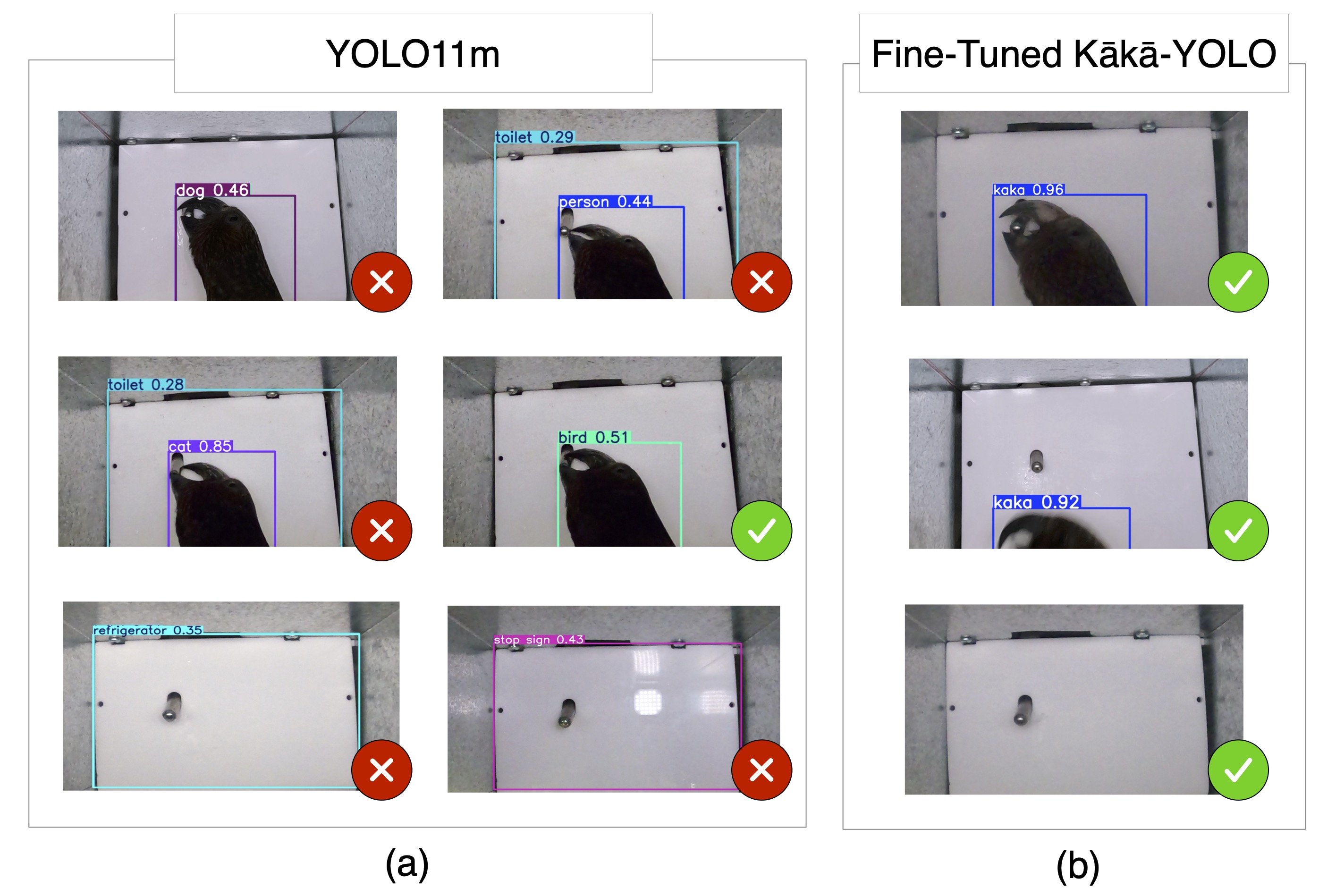}
}
\caption{(a) Comparison of the base YOLO11m model and (b) fine-tuned K\={a}k\={a}-YOLO model inference.}
\label{fig:yolo_errors}
%\vspace{-1em}
\end{figure}

\Cref{fig:yolo_model} illustrates our fine-tuning methodology using a subset of the unlabelled k\={a}k\={a} videos. To compare the performance of YOLO11m out-of-the-box with our fine-tuned model, we ran inference using a small selection of videos from the unlabelled datasets.  As shown in \cref{fig:yolo_errors} the base model often mislabels the bird and includes objects not of interest like the feeder, highlighting the importance of fine-tuning.

\subsubsection{Fine-Tuning Dataset}
To create a dataset for fine-tuning our YOLO model, we used a Grounding DINO model (tiny variant) from IDEA-Research \cite{liu2024}. 
Grounding DINO is a zero-shot object detection model based on the architecture used in other DINO variants, together with grounding pretraining (connecting visual data with text descriptions). The model detects objects based on text prompts, including objects unseen in its training data (i.e.\ zero-shot).

Adopting this approach enabled us to automate the creation of an annotated dataset by detecting a bird in the images and constructing bounding box coordinates. We randomly sampled 37 videos (30fps) of unlabelled k\={a}k\={a} birds recorded while collecting our first dataset at our feeding station. This dataset represented our lowest resolution images and least refined setup, allowing us to fine-tune our model on the most challenging data. This approach enabled us to develop a model that could generalise to our new, unseen datasets. 
 In previous studies, a common strategy was to extract frames at regular intervals, such as every 10th frame, which often ensured variations in the animal's position \cite{fintan2023,Xiao2023,Schofield2023}. Therefore, we adopted a similar strategy and extracted every 10th frame as shown in \cref{fig:yolo_model}a. We used Grounding DINO to detect a bird in the frame using the text prompt \texttt{"A bird."} with a detection threshold of 0.8 (\cref{fig:yolo_model}b). Through trial and error using a small set of three videos, we identified this threshold as effective. Higher values tended to capture frames with partial k\={a}k\={a}, which we considered insufficiently informative, while lower values failed to detect birds in more complex poses. 
For every successful detection, we generated normalised bounding box coordinates. 
We saved the image with the annotated bounding box (\cref{fig:yolo_model}c) then visually validated the images to confirm the successful object detection. We saved a copy of the image  (\cref{fig:yolo_model}d) and stored the coordinates in a labels text file (\cref{fig:yolo_model}e) for use in our fine-tuning dataset.  
The process generated 516 images with bounding box coordinates. 

\subsubsection{Fine-Tuning our YOLO Model} 
We chose a 70/30 train/test split, giving 361 images in our training dataset and 155 in the test set (\cref{fig:yolo_model}f). Our model contained one class, ``k\={a}k\={a}", and we trained\footnote{Training performed on an Apple Silicon MacBook M4 Max with 64GB memory and 40-core GPU using the MPS (Metal Performance Shaders) framework.} 
over 30 epochs 
with the recommended default parameters for our hardware and the YOLO model using a batch size of 16,
image size of 640, auto optimiser with a learning rate of 0.01 and momentum of 0.937. 
Through iterative experimentation, we found that training for 30 epochs was sufficient for model convergence, giving a final trained model 40.5MB in size.
We use common metrics in computer vision \cite{Padilla2020} to present the performance of our model, namely precision, recall, mAP@0.5, mAP@[0.5:0.95] :
\begin{itemize}
    \item Precision measures the proportion of correctly detected k\={a}k\={a} out of all detections made by the model.
    \item Recall measures the proportion of actual k\={a}k\={a} in the images that were correctly detected. 
    \item mAP@0.5 (mean Average Precision at an IoU threshold of 0.5) averages the precision when considering a bounding box is correct if it overlaps with a ground-truth box by at least 50\%.
    \item mAP@[0.5:0.95] (mean Average Precision averaged over multiple IoU thresholds from 0.5 to 0.95) provides a more comprehensive evaluation of the model's detection performance by averaging mAP scores across thresholds from 50\% to 95\% of overlap between predicted and ground-truth bounding boxes.
\end{itemize}
The final metrics for our model give a precision of  99.96\%, 
recall of 100\%, 
mAP@0.5 of 99.50\%, and
mAP@[0.5:0.95] of 99.49\%. 
These results indicate our fine-tuned model performs exceptionally well for our study.

\subsection{Frame Extraction}
\label{sec:frame_extraction}
\begin{figure}[h!]
\centerline{
%\vspace{-1em}
\includegraphics[width=0.95\textwidth]{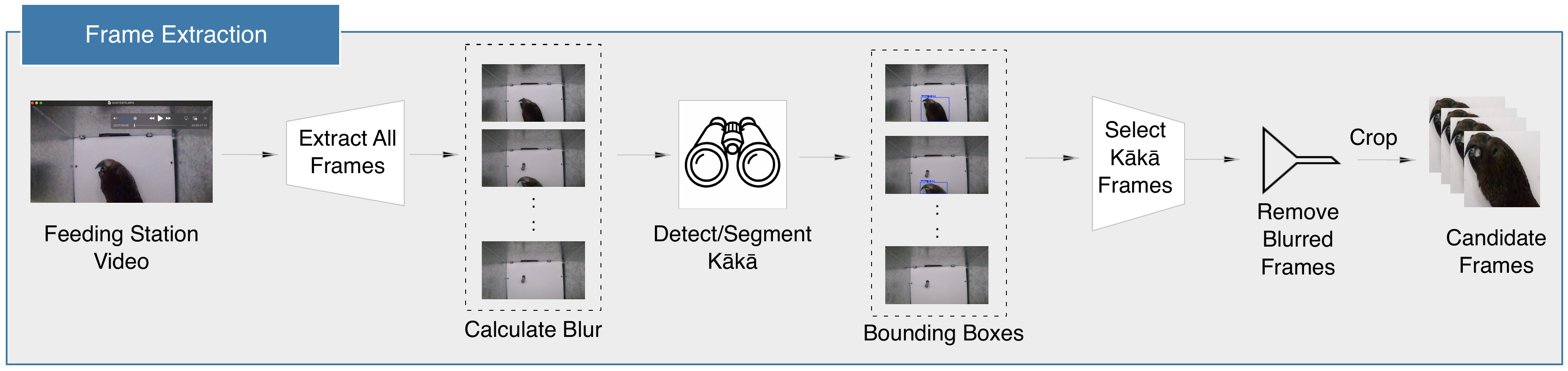}
}
\caption{Frame extraction stage of pipeline.}
\label{fig:frame_extraction}
%\vspace{-1em}
\end{figure}

Returning to our pipeline in \cref{fig:pipleline_overview}, the steps within the \textbf{Frame Extraction} stage are outlined in 
\cref{fig:frame_extraction}. 
We extracted all frames from each video in our three labelled datasets. For each frame, we calculated the Gunnar Farneb{\"a}ck motion score\footnote{Using the implementation from the Python OpenCV library.}. This step allowed us to identify blurred frames that may not be useful as key frames. The Gunnar Farneb{\"a}ck motion score estimates the motion of every pixel between a frame and its predecessor, referred to as dense optical flow. However, rather than directly comparing pixel values, it approximates the intensity of small local regions using a polynomial expansion function. It then quantifies the movement between frames by comparing the functions and computing their displacement \cite{Farnebaeck2003}. 

\begin{figure}[!h]
\centerline{
%\vspace{-1em}
\includegraphics[width=0.48\textwidth]{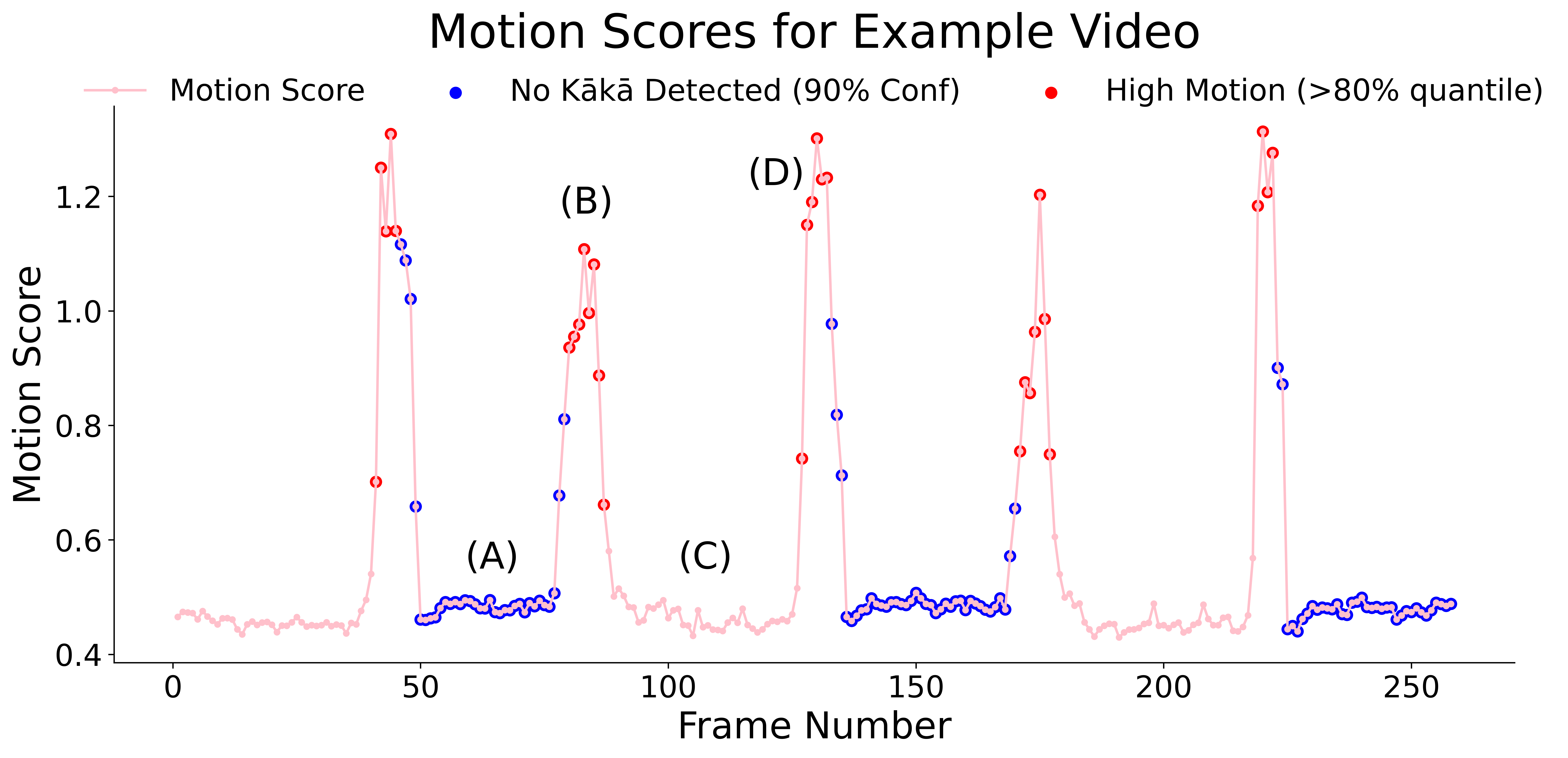}
}
\caption{Gunnar Farneb{\"a}ck motion scores (pink) for each frame using an example video
%in Dataset B
 indicating the amount of movement between consecutive frames. Blue points have no detected bird (A). 
Each peak following a blue set of low scored points represents the bird moving into the feeder (B). Consecutive pink points with low scores are the bird feeding (C). Each peak after a set of low scored pink points represents the bird leaving the feeder (D). Red points highlight frames considered high motion with blur. }
\label{fig:motion_scores}
%\vspace{-1em}
\end{figure}

We then applied our fine-tuned K\={a}k\={a}-YOLO model to detect k\={a}k\={a} in the frames (the segmentation element was not implemented during this preliminary study). Frames without detections or with high motion scores (indicating blur) were discarded. 
We used a trial-and-error approach to establish an acceptable threshold for identifying blurry frames.
 By analysing a small random sample of labelled videos, we found that removing the top 20\% of frames with the highest motion scores per video effectively reduced blurred images.   \Cref{fig:motion_scores} illustrates the motion scores for an example video 
 and shows how these scores reflect the bird's movement across video frames. We then cropped the remaining frames using bounding box coordinates generated by the object detection model, forming our candidate images for key frame selection.

\subsection{Key Frame Selection}
\label{sec:key_frame_selection}
\begin{figure}[!h]
\centerline{
%\vspace{-1em}
\includegraphics[width=0.95\textwidth]{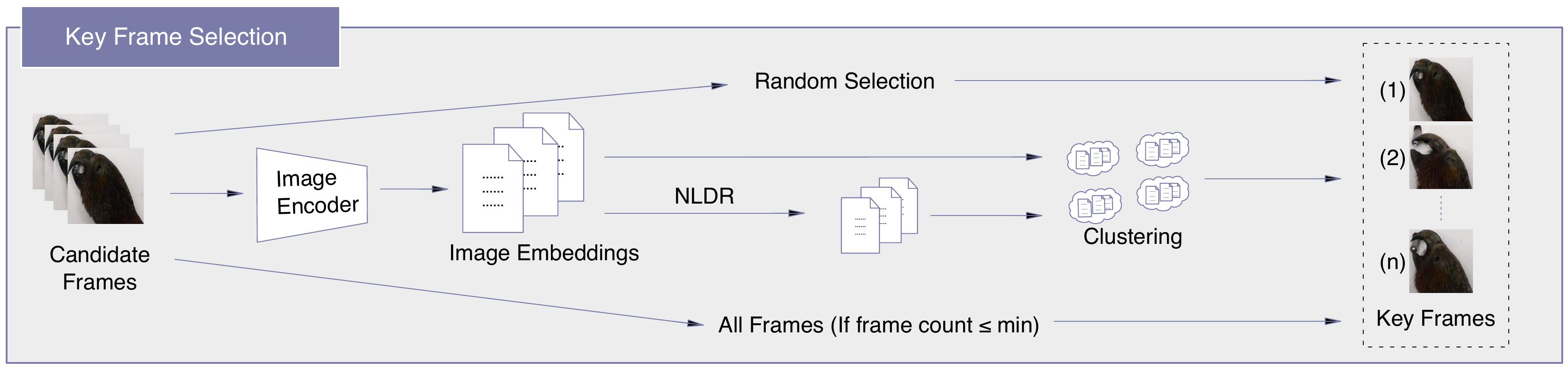}
}
\caption{Key frame selection stage of pipeline.}
\label{fig:key_frame_selection}
%\vspace{-1em}
\end{figure}

\Cref{fig:key_frame_selection} illustrates the steps in the \textbf{Key Frame Selection} stage of the pipeline in \cref{fig:pipleline_overview}(b). 
Following the creation of the candidate set of frames in the previous stage, we built embeddings of the frames using an image encoder.
An image embedding provides a unique numerical representation of an image,  capturing its important features and presenting the image in a format compatible with computer vision tasks.
We used a DINOv2 small model ViT-S \cite{DINOv22024} with an embedding dimension of 384 and patch size of 14 from the PyTorch Hub\footnote{https://pytorch.org/hub/}. This DINOv2 variant had demonstrated excellent performance in our previous study \cite{Maddigan2024}. 
The model is a self-supervised CV encoder developed by Meta AI and trained on unlabelled image data.  It creates image embeddings comparable to (weakly) supervised methods across various benchmarks without requiring fine-tuning, making it well-suited to our study based on unlabelled data.
We pre-processed each frame by resizing it to square (224x224) and using the standard ImageNet normalisation (mean=[0.485, 0.456, 0.406], std=[0.229, 0.224, 0.225]), ensuring DINOv2 processed the frames in the same way it learned during training. 

We considered two well-known unsupervised clustering methods to select a set of key frames -- \textit{k}-means \cite{macqueen1967} and \textit{k}-medoids \cite{Park2009}. 
The \textit{k}-means method groups similar data points into \textit{k} clusters by minimising the combined distance metric between each data point and the cluster centroid (mean value of all cluster points). It is an efficient iterative algorithm but is sensitive to outlier points.

The \textit{k}-medoids algorithm uses a similar iterative process, but instead of using centroids calculated from mean values, it uses actual points in the data (medoids) as cluster centres. 

One drawback with using these clustering methods on high-dimensional data such as image embeddings is they can suffer from the ``curse of dimensionality" \cite{Domingos2012} especially when the instances are sparse, as in our case where some videos produce few suitable frames to cluster. 

To address this issue, we experimented with applying a non-linear dimensionality reduction technique prior to clustering. We chose UMAP (Uniform Manifold Approximation and Projection) \cite{mcinnes2020}, a computationally efficient, state-of-the-art method that preserves both local and global structures and has been shown to improve clustering algorithms \cite{Allaoui2020,Healy2024}. \Cref{tab:umap_parameters} lists our UMAP parameter settings.

\begin{table}[htbp]
%\vspace{-1em}
\caption{Parameter settings for UMAP}
\begin{center}
\begin{tabularx}{\textwidth}{lX}
\hline
\noalign{\smallskip}
{Parameter}&
{Description} \vspace{2pt}\\
\hline
\noalign{\smallskip}
n\_components=5 & UMAP requires the output dimensionality to be lower than the number of instances to ensure stability of the lower-dimensional embedding. Exceeding this threshold results in a non-full rank matrix during the UMAP reduction algorithm.  This complication can lead to unreliable distance calculations and hinder manifold learning giving a poor preservation of local structure. For our datasets, some videos contained at most nine usable frames. Hence, to future-proof our method and allow for short video capture and media, we selected a dimension of 5.\\
min\_dist=0 & Setting the minimum distance hyper-parameter to zero makes points pack together densely, allowing for cleaner separations between clusters \cite{Allaoui2020}.\\
init=PCA & Principal Component Analysis (PCA) is a dimensionality reduction method to simplify complex data by mapping it into a lower-dimensional space and capturing the most variance in the data.  As it excels at retaining the global structure in the data, using PCA's first five principal components provides a good starting point for the UMAP initialisation \cite{Cristian2024}.\\
n\_neighbours=min(15,frame\_count) & We used the default of 15 but adjusted it to accommodate videos with fewer than 15 frames.\\
random\_state=42 & Using a seed ensures the reducibility of our experiments.
\vspace{2pt}\\
\hline
%\vspace{-2em}
\end{tabularx}
\label{tab:umap_parameters}
\end{center}
\end{table}

When using \textit{k}-means and \textit{k}-medoids, we experimented with $k\in\{5,6,7,\dots,20\}$ to identify clusters and select key frames representing each cluster. We aimed for at least five key frames, which would provide a good summary of the video, but not more than 20, as having too many would unnecessarily fill up storage space. 

For each value of \textit{k}, we calculated the silhouette score \cite{Rousseeuw1987} as defined in \cref{eqn:silhouette}. It measures the clustering quality by evaluating how well a data point fits within its assigned cluster compared to other clusters. A score close to 1 indicates good clustering, 0 indicates poorly defined clusters, and -1 indicates points may be allocated to wrong clusters. We chose the value of \textit{k} with the highest silhouette score.  

\begin{equation}
S = \frac{1}{n} \sum_{i=1}^{n}{\frac{b(i) - a(i)}{\max\{a(i), b(i)\}}}
\label{eqn:silhouette}
\end{equation}
\noindent where 
\( a(i) \) is the mean distance from point \( i \) to all other points in the \textit{same} cluster (intra-cluster distance) and 
 \( b(i) \) is the lowest mean distance from point \( i \) to points in the nearest \textit{different} cluster (nearest-cluster distance) in a sample of size \( n \).

Using these methods we chose one key frame to represent each cluster of frames.
For \textit{k}-means, we calculated the distance between each frame in the cluster and its centroid, and chose the frame closest to the centroid as the key frame. For \textit{k}-medoids, since the centroid is an actual frame, we used this frame as the key frame. 

To establish a baseline for comparison without clustering, we included two sets of randomly selected keyframes:
\begin{itemize}
    \item One set contained five key frames, the minimum number of frames we considered useful to represent a video.
    \item The other set contained seven keyframes, allowing us to assess the impact of increasing the number of key frames on the accuracy of our image matching.
    
\end{itemize}

Including random selection key frame sets enabled us to evaluate how well our proposed clustering methods performed. We also investigated whether using more key frames (in this case, seven instead of five) improved image matching accuracy. We chose not to include more than seven key frames because generating, processing, and storing a large dataset of images would be impractical for our pipeline and database. We considered seven frames to be realistic and relevant to our study goals. Media (short video or photos) with less than an set number of minimum candidate frames (e.g. $\leq 5$) will have all candidate frames selected as key frames.

In summary we explored six different methods to select a set of key frames:

\begin{enumerate}
    \item \textit{k}-means clustering
    \item \textit{k}-medoids clustering
    \item UMAP reduction followed by \textit{k}-means clustering
    \item UMAP reduction followed by \textit{k}-medoids clustering
    \item Random selection of five frames
    \item Random selection of seven frames
\end{enumerate}

\subsection{Similarity Matching}
\begin{figure}[!h]
\centerline{
%\vspace{-1em}
\includegraphics[width=0.95\textwidth]{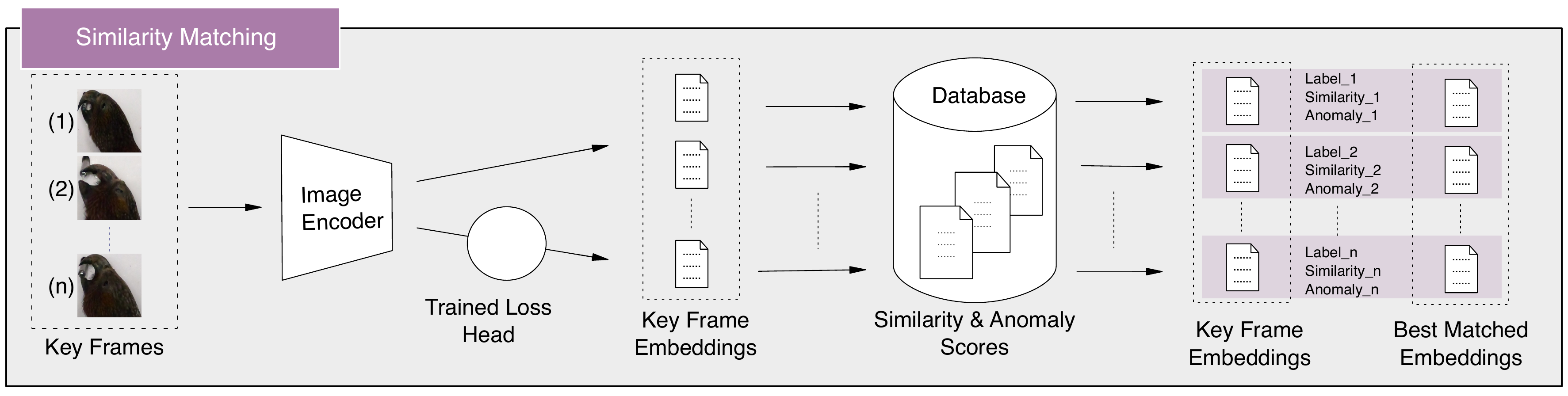}
}
\caption{Similarity matching stage of pipeline.}
\label{fig:similarity_matching}
%\vspace{-1em}
\end{figure}

\Cref{fig:similarity_matching} illustrates the steps in the \textbf{Similarity Matching} stage of the pipeline in \cref{fig:pipleline_overview}(c). The previous stage identified key frames to represent the video. This stage builds image embeddings of key frames for the purpose of similarity matching and finding the best matched bird in the database.  This stage also explores the efficacy of each key frame extraction method discussed during the previous stage.
In future work we will investigate training a loss head for the embedding encoders to determine their utility in improving the embedding space.  However, doing so will require a labelled dataset, which we intentionally avoided during this work to maintain an unsupervised approach

We evaluated three different embedding models (encoders) with different architectures and compared the quality of image embeddings by measuring the accuracy of our similarity matching process:
\begin{itemize}
    \item DINOv2: We continued to use the DINOv2 ViT-S model from our ``Key Frame Selection" stage (\cref{sec:key_frame_selection}) with embedding size 384.
    \item ResNet: This is an established well-known and widely-used model. We chose the ResNet50 model (specifically, ResNetV1.5) from Python’s TorchVision library\footnote{https://pytorch.org/vision/main/models/resnet.html}.
    This convolutional neural network (CNN) includes residual connections and is pre-trained on the IMAGENET1K\_V2 dataset. We removed the final classification layer, giving an image embedding size of 2048. 
    \item AIMv2: We used Apple’s Autoregressive Vision Transformer (ViT Large 224px) model \cite{fini2024}, a relatively recent and less adopted model in CV tasks.  Its inclusion in our study provided useful comparison against more established image encoders. We chose the AIM Python library TorchVision wrapper and applied pooling to the last layer to give an image embedding of size 1024.
\end{itemize}

We began by evaluating the performance of our three encoders on dataset A, which comprises our lowest-resolution videos and least refined setup. We chose this dataset as the basis for identifying the top-performing encoder as it presents the most challenging conditions in our video collection, allowing us to ensure the robustness, reliability, and performance of our chosen model. We then used this model for datasets B and C. To provide a baseline for comparison, we included the traditional hand-crafted SIFT/RANSAC approach explored in our previous study \cite{Maddigan2024}. In this method, SIFT detects keypoints and extracts feature descriptors, which are then matched between images. RANSAC subsequently removes incorrect matches due to noise.

We conducted similarity matching of the image embeddings using cosine similarity which is a widely used method for evaluating the similarity between high-dimensional embeddings. It measures the similarity between two vectors A and B by calculating the cosine of the angle between the vectors and determining whether the vectors are pointing in similar directions. 

The formula is given in \cref{eqn:cosine}.
\begin{equation}
\cos(\theta) = \frac{\mathbf{A} \cdot \mathbf{B}}{\|\mathbf{A}\| \|\mathbf{B}\|}
%    (A.B)/(|(|A|)|×||B||) 
\label{eqn:cosine}
\end{equation}
 To find the best match for each key frame we used \cref{eqn:cosine} to compare each frame embedding to all other frame embeddings in its dataset, excluding those extracted from the same video. Then, for each key frame embedding we assigned as the predicted label the leg band combination from the best matched frame embedding according to the highest similarity score. 

As an additional step we evaluated the performance of the image-matching method by comparing the predicted label of each key frame to its actual label.
If the labels matched, we considered that the bird in the image was correctly identified. Otherwise, we considered it incorrectly identified. 
We evaluated the image-matching accuracy of correctly identifying each labelled individual in the dataset and calculated the overall image-matching accuracy for each dataset. 
In this study we focused on closed-set re-ID hence we omitted the anomaly scoring at this stage.

\subsection{Evaluation}
\begin{figure}[h!]
\centerline{
%\vspace{-1em}
\includegraphics[width=0.95\textwidth]{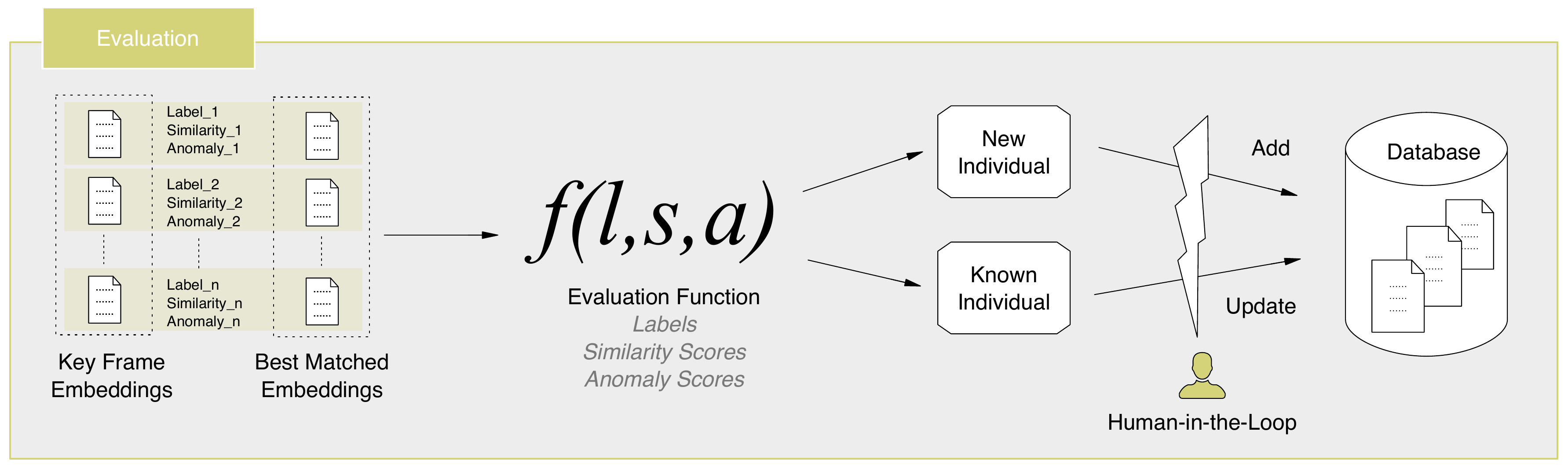}
}
\caption{Evaluation stage of pipeline.}
\label{fig:evaluation}
%\vspace{-1em}
\end{figure}

In the previous stage, for each of the key frames we selected a best matched frame from the dataset using the highest similarity score, and noted its label. \Cref{fig:evaluation} shows the flow of the \textbf{Evaluation} stage of the pipeline in \cref{fig:pipleline_overview}(d), taking the key frame embeddings, matched labels, and scores as input. We built a simple function \textbf{\textit{f}} to take these values and identify the individual in the video. As this study used a closed-set dataset all individuals were known hence, anomaly scores were not calculated, the function did not  evaluate if the individual was a new bird or not, and the human-in-the-loop interrupt was not required. 

We implemented two efficient and straightforward variations of the evaluation function \textbf{\textit{f}}:

\begin{enumerate}
    \item Threshold-based approach: We applied a threshold of 60\% or 80\%, where an individual in a video was considered correctly identified if at least this percentage of its frames matched the same bird.
    \item Majority voting: We used a simple majority vote, where the bird that appeared in the most frames from a video we deemed to be the  individual present in the video.
\end{enumerate}

We evaluated the accuracy of identifying individual k\={a}k\={a} in videos using both variations of the evaluation function \textbf{\textit{f}}. This allowed direct comparison with our earlier heuristic-based key frame selection approach in our previous work \cite{Maddigan2024}. 

We used statistical significance tests to evaluate the performance of our proposed video re-ID methods. Specifically, we compared each proposed extraction method with every other proposed method and our previous heuristic method. This approach allowed us to identify the top-performing methods and determine if our new approaches improve upon our earlier study.
We chose the McNemar statistical test of homogeneity\cite{McNemar1947}, a non-parametric test that does not require normally distributed data. We applied a Bonferonni correction for multiple contrasts. The McNemar test is suitable for detecting differences between paired samples on a dependent categorical variable with only two categories (dichotomous). 
In our study, the samples are our videos, and the dichotomous variable represents success or failure in identifying an individual bird in a video. The test measures the difference in the count of errors between the two selected methods, with our null hypothesis stating both models have a similar proportion of errors. 
The McNemar test\footnote{We chose the binomial distribution implementation from the Python statsmodels library.} is equivalent to the t-test but is used for binary targets instead of continuous ones. It is considered the most appropriate test for comparing dichotomous outcomes using two dependent samples \cite{Fisher2011}. Using this test, we determined whether our proposed methods demonstrated statistically significant improvements over our previous heuristic method and identified the most effective approaches.

\subsection{Similarity Matching Against a Partially-Labelled Dataset}

To assess the accuracy of image matching from an alternative perspective, we conducted a further experiment using the images in Dataset A generated via the heuristic-based method.
We compared each image in the labelled dataset against a partially-labelled dataset comprising labelled and unlabelled images (5,977 images).  

We used two methodologies -- SIFT and DINOv2 embeddings.  We then compared them against two methods from prior research which used a \textit{k}-means image masking method -- SIFT with background masking and SIFT with combined background and nozzle masking  \cite{fintan2023}.

\section{Results and Discussion}

In the following sections we present our results from the four stages in our pipeline depicted in \cref{fig:pipleline_overview} -- \textbf{Frame Extraction} (\cref{fig:frame_extraction}), \textbf{Key Frame Selection} (\cref{fig:key_frame_selection}), \textbf{Similarity Matching} (\cref{fig:similarity_matching}), and \textbf{Evaluation} (\cref{fig:evaluation}). 

\subsection{Frame Extraction}
For each of our datasets, \cref{tab:frame_summary} gives a count of the number of labelled birds, the number of videos, the total frames extracted from all  videos in the dataset, the number of frames with k\={a}k\={a} detected, and the final count of candidate key frames following removal of high motion images. 
(Breakdowns of these figures by bird label within each dataset is provided in \cref{tab:ds1_detail,tab:ds2_detail,tab:ds3_detail} of Appendix \ref{appx:ds_summaries}.)

\begin{table}[htbp]
%\vspace{-1em}
\caption{Dataset Summary from the Frame Extraction Stage}
\begin{center}
\begin{tabular}{crrrrr}
\hline 
\noalign{\smallskip}
{Dataset} & {Labelled Birds}& 
{Videos}&{Total Frames}
&{K\={a}k\={a} Frames}&{Low Motion Frames}\vspace{2pt}\\
\hline \noalign{\smallskip}
A & 7 & 128	& 64,791	&35,825	& 28,536\\
B & 15 & 1,821	& 234,911 &137,106	&107,759\\
C & 14 & 517 & 504,247 &231,350 &184,551\\\noalign{\smallskip}
\hline
%\vspace{-2em}
\end{tabular}
\label{tab:frame_summary}
\end{center}
\end{table}

\subsection{Key Frame Selection}

\subsubsection{Clustering of Frames}
To illustrate our approach, we continue with the example video introduced in  \cref{fig:motion_scores} with 259 frames. % Video GH014677
We detected k\={a}k\={a} in 151 of these frames, and after removing 31 frames which we considered high-motion images, our candidate set for key frame selection contained 120 frames. 
We compared the performance of \textit{k}-means and \textit{k}-medoids clustering algorithms, both with and without prior dimensionality reduction using UMAP. Our results showed that \textit{k}-means with UMAP reduction yields 5 clusters based on the highest silhouette score, whereas \textit{k}-means, \textit{k}-medoids, and \textit{k}-medoids with UMAP reduction result in 6 clusters.

\begin{figure}[!h]
\centerline{
%\vspace{-1em}
\includegraphics[width=3.6in]{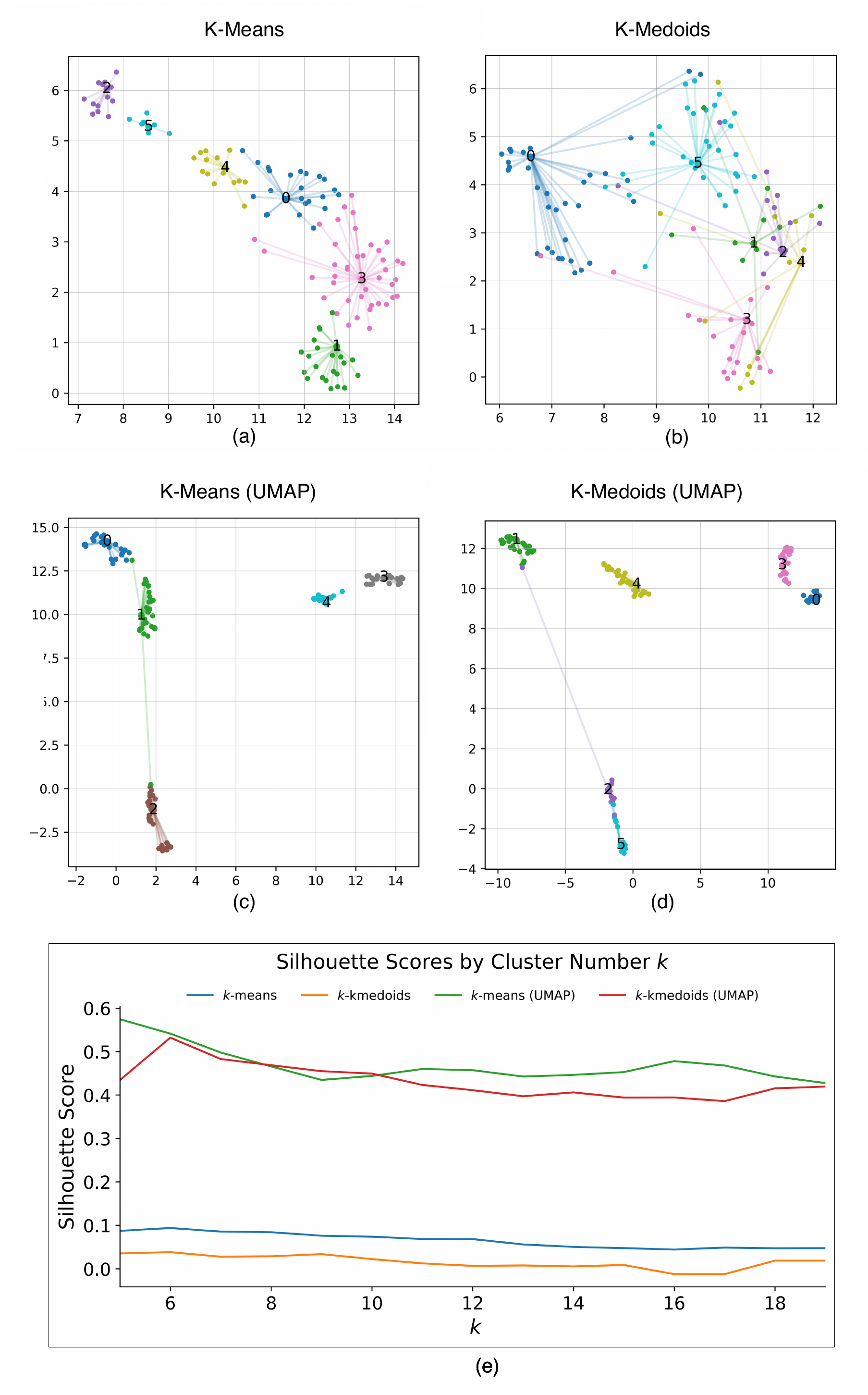}
}
\caption{Visualisation of 120 image embeddings using UMAP for the selection of key frames from an example video. % in dataset B. 
(a) \textit{k}-means clustering with 6 clusters, 
(b) \textit{k}-medoids clustering with 6 clusters, (c) \textit{k}-means clustering of UMAP reduced embeddings with 5 clusters,
(d) \textit{k}-medoids clustering of UMAP reduced embeddings with 6 clusters, 
(e) silhouette scores by \textit{k} for each method.}
\label{fig:custering_eg}
%\vspace{-1em}
\end{figure}

We used UMAP for NLDR of the embeddings prior to clustering. However, UMAP is also a useful tool for reducing embeddings to two dimensions for visualisation tasks.  Hence we visualise the results by 
reducing the embeddings to two dimensions using UMAP, as pictured in \cref{fig:custering_eg}, with each cluster assigned a distinct colour. 
The \textit{k}-means clusters (\cref{fig:custering_eg}a) appear more defined than those from \textit{k}-medoids (\cref{fig:custering_eg}b). It is important to note that some information will be lost through the UMAP reduction process, with only essential features preserved in the reduced embeddings. Therefore, the visualisations of the UMAP versions of \textit{k}-means (\cref{fig:custering_eg}c) and \textit{k}-medoids (\cref{fig:custering_eg}d)  may give the illusion of superior clustering compared to the original embeddings (\cref{fig:custering_eg}a-b). Nevertheless, these visualisations still offer a valuable means of comparing the relative performance of the \textit{k}-means and \textit{k}-medoids algorithms using UMAP reduced embeddings. \Cref{fig:custering_eg}e illustrates the silhouette scores across increasing values of \textit{k}, showing peak values at 5 clusters for \textit{k}-means and with 6 clusters for the remaining three methods.  

\subsubsection{Silhouette Scores}

\begin{figure}[!h]
\centerline{
%\vspace{-1em}
%\includegraphics[width=3.6in]{silhouette.pdf}
%\includegraphics[width=0.48\textwidth]{silhouette_scores_box.png}
\includegraphics[width=0.98\textwidth]{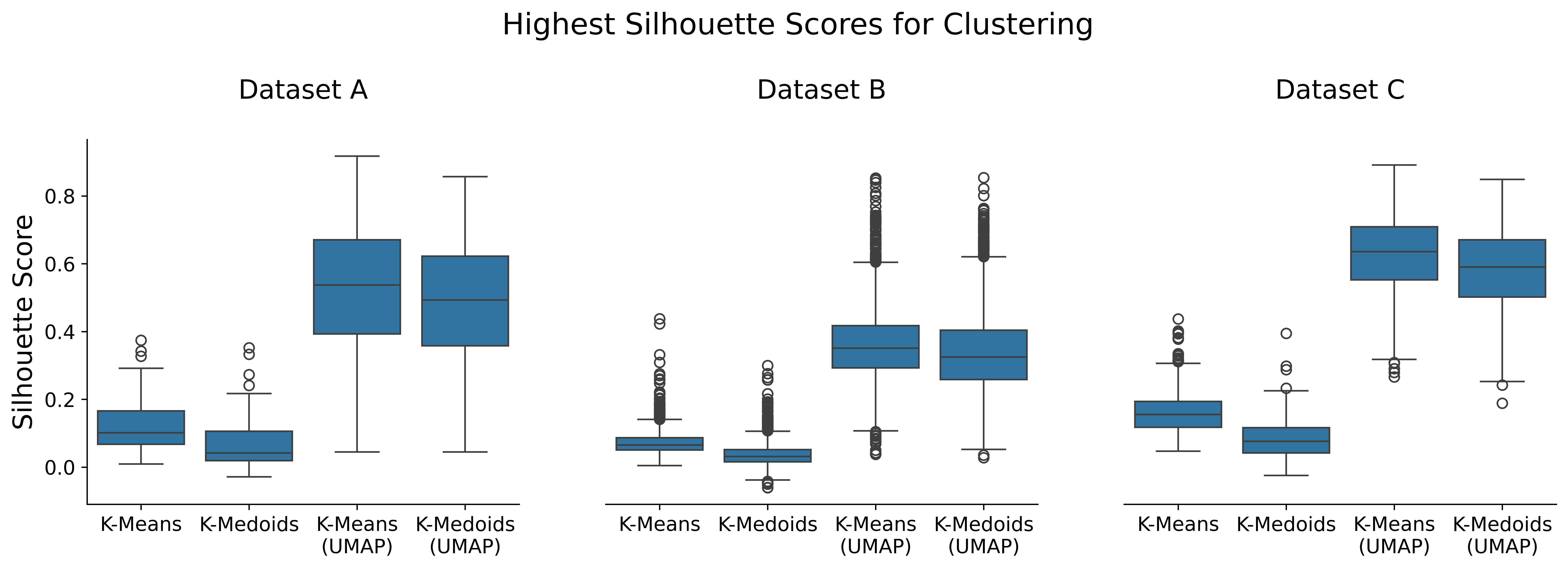}
}
\caption{Highest silhouette scores for videos in datasets A, B, and C using \textit{k}-means and \textit{k}-medoids clustering with and without UMAP reduction.}
\label{fig:silhouette}
%\vspace{-1em}
\end{figure}

To determine the most suitable number of key frames $k$ (clusters) for each video, we selected the value of $k$, giving the highest silhouette score for $k$ between 5 and 20. \cref{fig:silhouette} displays the distribution of these scores calculated for each video across our three datasets.

Visually we observe two trends -- (1) \textit{k}-means clustering tends to produce higher scores than \textit{k}-medoids, and (2) applying UMAP before clustering results in higher scores. 
It is important to note that clustering based on the reduced UMAP embeddings effectively alters the underlying data representation and computed distance values. As a result, the silhouette scores are not directly comparable to those of the original embeddings. Nevertheless, most scores are above zero, suggesting the clustering process retains some meaningful structure.

\subsubsection{Dataset Key Frames}
\cref{tab:key_frame_summary} summarises the key frame counts for each dataset using our various selection methods.  For comparison, we have included the image counts from our previous study's heuristic method \cite{Maddigan2024}. 
The results from our proposed methods show that across all datasets, 
the ``Random 5'' selection yields the \textit{smallest} number of key frames, ``\textit{k}-means (UMAP)'' generates the \textit{largest} number of key frames for datasets A and B and ``\textit{k}-medoids'' the \textit{largeest} number for dataset C.

Comparing our proposed methods to our previous heuristic-based method, we find that the frame counts are generally comparable in datasets A and C. However, in dataset B, our new methods generate 3-6 times more frames. 
We attribute this discrepancy to the different frame selection rates during the heuristic-based work. During that study, on average, we selected 6-7 frames per video in datasets A and C (with 128 and 517 videos, respectively). However, dataset B had a notably larger quantity of videos (1821), and we heuristically selected only 1-2 frames per video.

\begin{table}[htbp]
%\vspace{-1em}
\caption{Count of Key Frames by Dataset and Selection Method}
\begin{center}
\begin{tabular}{lrrr}
\hline
\noalign{\smallskip}
{Selection Method}& 
{Dataset A} & 
{Dataset B} &
{Dataset C}\vspace{2pt}\\
\hline
\noalign{\smallskip}
\textit{k}-means & 816 & 13,502 & 4,221\\
\textit{k}-medoids & 921 & 13,866 & 4,720\\
\textit{k}-means (UMAP) & 1,590 & 19,362 & 4,006 \\
\textit{k}-medoids (UMAP) & 962 & 12,118 & 4,056\\
Random 5 & 640 & 9,105 & 2,585\\
Random 7 & 895 & 12,747 & 3,619\\
Heuristic \cite{Maddigan2024}  & 795 & 2,910 & 2,996\\
\noalign{\smallskip}
\hline
%\vspace{-2em}
\end{tabular}
\label{tab:key_frame_summary}
\end{center}
\end{table}

\subsection{Similarity Matching}

\subsubsection{Image Embeddings of Key Frames}
\begin{figure}[!h]
\centerline{
%\vspace{-1em}
\includegraphics[width=3.6in]{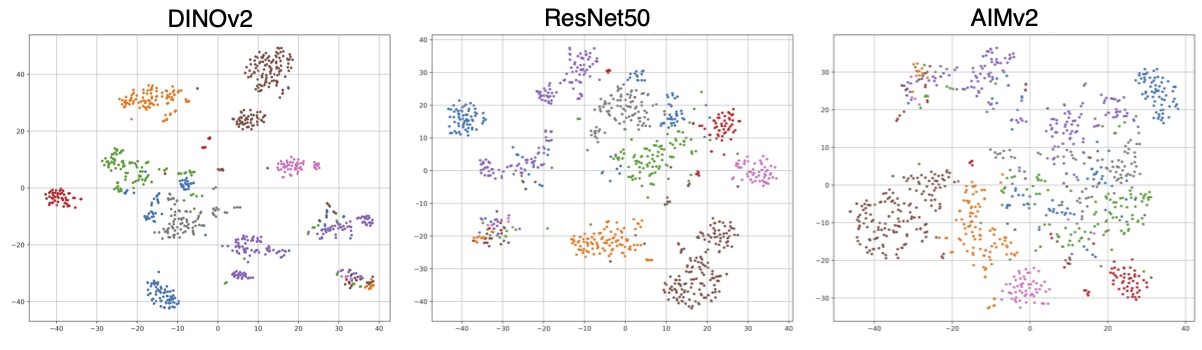}
}
\caption{Dataset A image embeddings for key frames from \textit{k}-means clustering using DINOv2, ResNet50, and AIMv2.}
\label{fig:emb_kmeans}
%\vspace{-1em}
\end{figure}

\begin{figure}[!h]
\centerline{
%\vspace{-1em}
\includegraphics[width=3.6in]{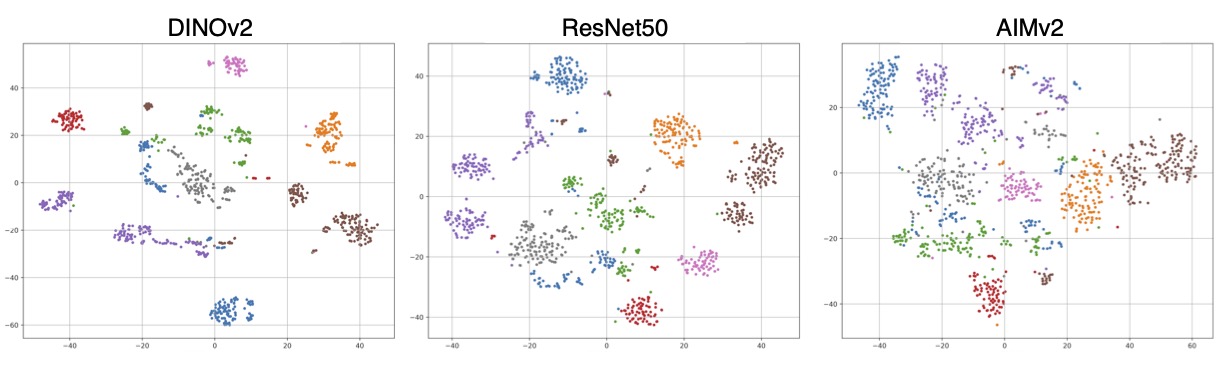}
}
\caption{Dataset A image embeddings for for key frames from \textit{k}-medoids clustering using DINOv2, ResNet50, and AIMv2.}
\label{fig:emb_kmedoids}
%\vspace{-1em}
\end{figure}

We created image embeddings for our datasets of keyframes using three different encoders -- DINOv2, ResNet50, and AIMv2. To establish the quality of the embeddings, we visualised the results for dataset A by reducing the embeddings to 2-D using UMAP. The embeddings produced by \textit{k}-means clustering are shown in \cref{fig:emb_kmeans} while those from \textit{k}-medoids are presented in  \cref{fig:emb_kmedoids}. Labelled data is not used in clustering but for visual clarity we assign a distinct colour to each labelled bird.  From these figures, we can make two key observations -- 
(1) \textit{k}-medoids may tend to produce marginally more separated and defined clusters for each bird label compared to its \textit{k}-means counterparts (despite weaker silhouette scores), and 
(2) among the three encoders, DINOv2 appears to create the best embeddings, while AIMv2 performs the worst.  Given DINOv2's superior performance, we focused on this model for further analysis. 

\begin{figure}[!h]
\centerline{
%\vspace{-1em}
\includegraphics[width=2.6in]{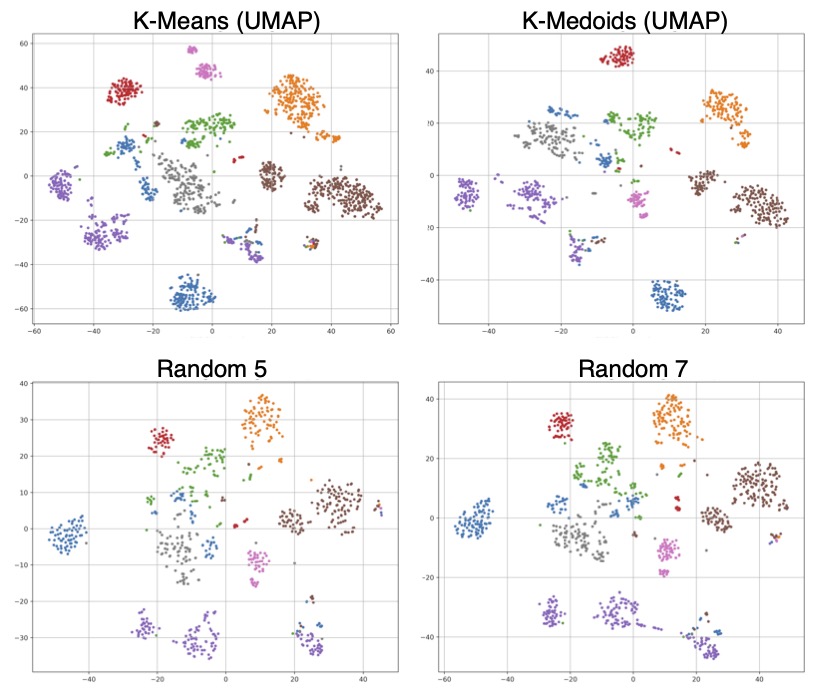}
}
\caption{DINOv2 Image embeddings for Dataset A key frames from UMAP reduced \textit{k}-means and \textit{k}-medoids clustering  (top row), and Random 5 and Random 7 (bottom row).}
\label{fig:emb_others}
%\vspace{-1em}
\end{figure}

\Cref{fig:emb_others} visualises the embeddings of \textit{k}-means and \textit{k}-medoids after applying UMAP reductions and the Random 5 and Random 7 selections of key frames. 
The results suggest that during the key frame selection step applying UMAP may not significantly impact clustering quality as the embeddings appear neither better nor worse than their non-UMAP counterparts. 
%In addition, visually \textit{k}-medoids appears superior to \textit{k}-means although this is not reflected in the overall silhouette scores. 
Furthermore, even random selection produces relatively well-defined embeddings. 
(Visualisations of the DINOv2 embeddings for datasets B and C are given in \cref{fig:emb_ds2} and \cref{fig:emb_ds3} of Appendix \ref{appx:emb}). 

Therefore, from a visual inspection of the embeddings, which is not strongly supported by the silhouette scores, we find there is little evidence to support a difference between  \textit{k}-means,  \textit{k}-medoids, their UMAP counterparts, and random selection. We address this further in \Cref{sec:stat_test} with results of statistical testing.

\subsubsection{Accuracy of Image Matching}
The accuracy of our similarity matching in identifying individual k\={a}k\={a} within images are presented in \cref{tab:similarity_results_images}. We evaluated three image encoders on dataset A and the traditional SIFT/RANSAC approach. For datasets B and C, we used DINOv2 only, which is our best-performing encoder from dataset A. For comparison, we included metrics from our heuristic key frame selection method used during our previous study. The most accurate algorithm for each dataset is highlighted in bold font. For dataset A, the DINOv2 encoder used on key frames selected via \textit{k}-medoids clustering on the reduced UMAP embeddings achieved the highest accuracy of 96.2\%. Continuing with the DINOv2 encoder, the \textit{k}-medoids algorithm using the original image embeddings emerged as the top performer for datasets B and C both with 98.6\% accuracy. Notably, the performance difference over all proposed methods was relatively marginal.

\begin{table*}[htbp]
%\vspace{-1em}
\caption{Accuracy of Image Matching}
\begin{center}
\begin{tabular}{clccccccc}
\hline
\noalign{\smallskip}
{Dataset}&
{Encoder} &
{\textit{k}-means}&
{\textit{k}-medoids}&
{\textit{k}-means}&
{\textit{k}-medoids}&
{Random}&
{Random} &
{Heuristic}\vspace{2pt}\\
&&&&(NLDR)&(NLDR)&5&7&\\
\hline
\noalign{\smallskip}
A& DINOv2 & 93.9 &	92.9 &	95.5 &	\textbf{96.2} &	94.5&	95.3 	 & 93.8 \\
& ResNet50 & 89.0 & 92.2 & 93.5  &	93.2 &	90.5 &	 91.1	& 90.8\\
& AIMv2 & 84.4 &	86.9 & 89.6 &	88.8 &	 	87.2	& 87.9 & 81.9 \\
& SIFT & 81.0 & 87.4 &	81.9 &	88.9 &	86.3 & 87.3 & 85.9\vspace{2pt} \\
\hline
\noalign{\smallskip}
B&DINOv2&97.8 &	\textbf{98.6}&	98.5&	98.4&	97.8&	98.0 & 97.8\vspace{2pt}\\
%\hline
%\noalign{\smallskip}
C & DINOv2 & 96.4& \textbf{98.6} & 97.6 & 97.2 & 	96.6& 96.9 & 97.6 \\
\noalign{\smallskip}
\hline
%\vspace{-2em}
\end{tabular}
\label{tab:similarity_results_images}
\end{center}
\end{table*}

\subsection{Evaluation}
The accuracy of our methods in identifying individual k\={a}k\={a} within videos using results from the DINOv2 image matching is presented in \cref{tab:similarity_results_videos}. 
Results are reported for the evaluation function \textbf{\textit{f}} using the threshold approach and majority voting.
We highlight the highest accuracy for each dataset in bold font. 
Across all datasets, voting outperforms our threshold approach. 
The accuracy of key frame matching shown previously in \Cref{tab:similarity_results_images} showed little variation across methods and hence to identify an individual in a video \cref{tab:similarity_results_videos} shows the random selection methods are competitive with our clustering-based approaches, with each improving upon the previous heuristic-based selection method. 

\begin{table*}[htbp]
%\vspace{-1em}
\caption{Accuracy of Re-Identification in Videos using DINOv2}
\begin{center}
\begin{tabular}{clccccccc}
\hline
\noalign{\smallskip}
{Dataset}&
{Eval.} &
{\textit{k}-means}&
{\textit{k}-medoids}&
{\textit{k}-means}&
{\textit{k}-medoids}&
{Random}&
{Random} &
{Heuristic}\vspace{2pt}\\
&Fn. \textbf{\textit{f}}&&&(UMAP)&(UMAP)&5&7&\\
\hline
\noalign{\smallskip}
A&60\%&96.1&	95.3& 95.3&	98.4&\textbf{98.4}&97.7&	95.3\\
&80\%& 93.0	&94.5& 93.8 & 94.5 &94.5&	94.5 	&90.6\\
&Voting&97.7 	&96.1&	97.7 &	\textbf{98.4}&\textbf{98.4}&	\textbf{98.4}&95.3\vspace{2pt}\\
\hline
\noalign{\smallskip}
B&60\% &	98.6	&99.0& 99.0&	98.8	&99.0&	98.7 &97.4\\
&80\%& 97.3	& 98.5 &97.7 &	98.2&98.1& 97.1 &97.3\\
&Voting& \textbf{99.3}	 &	\textbf{99.3}&\textbf{99.3}& 99.2		&99.2&\textbf{99.3}&97.9 	\vspace{2pt}\\
\hline
\noalign{\smallskip}
C&60\% 	& 97.7	&98.6& 98.6 & 	98.3 &98.6&97.9&97.5\\
&80\%& 	95.7 &98.1	& 97.1 &	97.3&97.1&  96.7&95.6\\
&Voting 	&98.6 &	\textbf{98.8}	& \textbf{98.8} &\textbf{98.8} 	& 	98.6&\textbf{98.8}&98.3\\
\noalign{\smallskip}
\hline
%\vspace{-2em}
\end{tabular}
\label{tab:similarity_results_videos}
\end{center}
\end{table*}

When evaluating the performance of different methods to select key frames from videos, we must balance the trade-off between the number of key frames produced and the accuracy achieved. For example, in Dataset A, applying \textit{k}-means to UMAP-reduced embeddings (1,590 images) generated nearly twice as many frames as using \textit{k}-means without UMAP reduction (816). However, the video accuracy was the same for both methods (97.7\%)

\subsection{Similarity Matching Against a Partially-Labelled Dataset}

The image matching results using the partially-labelled dataset are presented in \cref{table:results_compare}. Our findings indicate that SIFT without masking achieves an overall accuracy of 65.3\%, 
which improves to 70.6\% by including a background mask, and further increases to 78.8\% with both background and nozzle masks. 

DINOv2 outperforms SIFT in the absence of masking (69.9\% vs 65.3\%). Yet, when introducing background-only and background-nozzle masks, SIFT outperforms DINOv2. However, the masking used with SIFT in our the prior research was custom-designed and not easily transferable across datasets, limiting its future application.  
In this experiment, we did not explore the application of segmentation masks with DINOv2 to determine if it improved the performance in line with that of SIFT masking. 

\begin{table}[!h]
\centering
\caption{Comparative Analysis of Model Results With and Without Masks for Partially-Labelled Dataset}
\label{table:results_compare}
\resizebox{\columnwidth}{!}{%
\begin{tabular}{@{}lrrrrrrrrr@{}}
\toprule
 & 
  \multicolumn{1}{l}{} &
  \multicolumn{2}{c}{(A) SIFT} &
  \multicolumn{2}{c}{(B) SIFT} &
  \multicolumn{2}{c}{(C) SIFT} &
  \multicolumn{2}{c}{(D) DINO} \\
   & 
  \multicolumn{1}{l}{} &
  \multicolumn{2}{c}{Background + Nozzle} &
  \multicolumn{2}{c}{Background Only} &
  \multicolumn{2}{c}{No Mask} &
  \multicolumn{2}{c}{No Mask} \\
Label &
  \multicolumn{1}{l}{Total} &
  \multicolumn{1}{l}{Correct} &
  \multicolumn{1}{r}{Accuracy} &
  \multicolumn{1}{l}{Correct} &
  \multicolumn{1}{r}{Accuracy} &
  \multicolumn{1}{l}{Correct} &
  \multicolumn{1}{r}{Accuracy} &
  \multicolumn{1}{l}{Correct} &
  \multicolumn{1}{r}{Accuracy} \\ \midrule
L-MB  & \multicolumn{1}{r|}{139} & 96  & \multicolumn{1}{r|}{0.691} & 79 & \multicolumn{1}{r|}{0.568} & 80 & \multicolumn{1}{r|}{0.576} & 97&0.698 \\
LM-G  & \multicolumn{1}{r|}{102} & 65  & \multicolumn{1}{r|}{0.637} & 54  & \multicolumn{1}{r|}{0.529} &  46 & \multicolumn{1}{r|}{0.451}  & 35&0.343\\
O-RS  & \multicolumn{1}{r|}{202} & 191 & \multicolumn{1}{r|}{0.946} & 187 & \multicolumn{1}{r|}{0.926} &  168 & \multicolumn{1}{r|}{0.832}  & 181&0.896\\
MR-R  & \multicolumn{1}{r|}{57}  & 49  & \multicolumn{1}{r|}{0.860} & 45  & \multicolumn{1}{r|}{0.790} & 40  & \multicolumn{1}{r|}{0.702} & 41&0.719 \\
WS-P  & \multicolumn{1}{r|}{139} & 86  & \multicolumn{1}{r|}{0.619} & 81 & \multicolumn{1}{r|}{0.583} & 81 & \multicolumn{1}{r|}{0.583}  & 105&0.755\\
Y-GM  & \multicolumn{1}{r|}{9}   & 8   & \multicolumn{1}{r|}{0.889} & 7   & \multicolumn{1}{r|}{0.778} & 7  & \multicolumn{1}{r|}{0.778}  & 7&0.778\\
YM-Y  & \multicolumn{1}{r|}{147} & 128 & \multicolumn{1}{r|}{0.871} & 113 & \multicolumn{1}{r|}{0.769} & 95 & \multicolumn{1}{r|}{0.646} & 104&0.707 \\ \midrule
Total & 795 & 623 & 0.788 & 566 & 0.706 & 517 & 0.653 & 570 & 0.699\\ \bottomrule
\end{tabular}%
}
\end{table}

\subsection{Statistical Testing}
\label{sec:stat_test}
We calculated the p-values from McNemar's statistical significance tests with a Bonferonni correction for our video matching using the evaluation function \textbf{\textit{f}} with majority voting.  The results show that all our proposed key frame extraction methods give p-values greater than 0.05, indicating insufficient evidence to reject the null hypothesis. Therefore, we conclude that all methods perform equally well with a similar error rate. 

However, we observe different results when comparing our proposed methods to the heuristic-based method in our previous work, but only for dataset B. Here, we observed p-values less than 0.05 for all our proposed methods, suggesting significant differences in performance. As noted earlier, only 1-2 key frames per video were selected during the heuristic-based method for dataset B and hence re-ID for videos in dataset B was significantly less accurate using that approach.  

Our results suggest that the choice of clustering algorithm and using a dimensionality reduction technique has a limited influence on re-ID accuracy for our datasets, as the difference in performance is small compared to our random methods, and considered statistically non-significant. One potential disadvantage of random selection is that it may fail to capture all bird poses within a video, and hence, clustering may mitigate this. However, in our controlled environment, we minimised variations in bird positioning through our setup, reducing the impact of this occurring. In future, extending our study to include trail camera footage will help determine the effectiveness of clustering-based approaches in more diverse environments.

\subsection{Limitations}
While our study contributes valuable research to the field of wildlife re-ID, specifically k\={a}k\={a}, we acknowledge several limitations. We collected our datasets in a controlled environment with minimal occlusions, good lighting, and high-quality cameras.
The footage was captured over 15 months, with each dataset spanning 1 to 3 months. Dividing the recordings into three datasets avoided temporal variations within each dataset and changes that may have occurred in birds spanning multiple datasets. Another limitation is that we solely used UMAP for dimensionality reduction and \textit{k}-means and \textit{k}-meloids for clustering. Although we attempted to use HDBSCAN \cite{Campello2013}, it was not effective due to the somewhat small number of instances (k\={a}k\={a} frames) in some videos, and in addition, there was limited control of cluster size.

\subsection{Future Work}

Future work will entail expanding our datasets by collecting trail media to address the limitations of our existing datasets and building on the concepts presented in this study. Removing high motion frames from our set of candidate key frames was essential in improving the accuracy of k\={a}k\={a} video re-ID, however, we considered the Gunner-Farnebäck method was computationally slow. Future work will focus on developing a faster, customised solution to address this limitation.

By including a fine-tuned object detection model in our pipeline, we tailored our approach to be species-specific. As we collect more trail images, we plan to expand the model training and fine-tune it to enhance its performance and robustness across various settings. To further strengthen our pipeline, we intend to extend this component into a fine-tuned segmentation model to remove all additional background noise. In addition, we aim to broaden the applicability of our methodology by including other similar species, such as the kea or k\={a}k\={a}p\={o}. Our modular pipeline design allows for seamless swapping of components, enabling easy integration of other fine-tuned models for detecting different species, making it adaptable and scalable for future applications.

The choice of encoder model for generating image embeddings is important, as some architectures are better suited to this task and our specific datasets than others.  However, we did not fine-tune the encoder heads to optimise the embedding spaces in this study. Doing so would have required a labelled dataset, which we intentionally avoided using to maintain an unsupervised approach and retain our three datasets for evaluation. However, fine-tuning with a labelled dataset could further improve our encoders' performance and will be explored in later studies.

In future work we anticipate lower accuracy rates with the addition of trail recordings, providing a valuable opportunity to evaluate and refine our methodology, including alternative embedding encoders, model fine-tuning, clustering methods, and dimensionality reduction techniques. 

\section{Conclusion}
In this study we presented an automated AI-based pipeline for extracting key frames from videos to improve the accuracy of re-ID of individual k\={a}k\={a}.
Our findings show that using a fine-tuned AI object detection model to identify frames containing k\={a}k\={a},  cropping the images to minimise background noise, employing CV techniques to remove high-motion frames, then selecting key frames through clustering (or random selection) produces an image dataset rich in features and suitable for recognising individual k\={a}k\={a} with high accuracy.

While key frame extraction has been explored in various domains, its application to wildlife re-ID tasks is limited. Our k\={a}k\={a} case study is an important contribution to on-going research in this field. Using AI and computer vision, we aim to continually improve the recognition of k\={a}k\={a} individuals, thereby reducing reliance on traditional methods such as leg banding and manual tracking. These conventional approaches are invasive, time-consuming, and error-prone, making our automated AI-based pipeline a more effective and desirable alternative for re-ID and hence monitoring and managing wildlife populations such as k\={a}k\={a}.
\newpage
\appendices
\section{Dataset Summaries from the Frame Extraction Stage}
\label{appx:ds_summaries}
\begin{table}[htbp]
\caption{Results for Dataset A}
\begin{center}
\begin{tabular}{lrrrr}
\hline
\textbf{Band} & \textbf{{Videos}}& 
\textbf{{Total Frames}}&\textbf{{K\=ak\=a} Frames} &
\textbf{Low Motion Frames} \\
\hline
L-MB & 18 & 13,803 & 7,130 & 5,687\\ 
LM-G & 16 & 9,080 &4,686 & 3,734  \\ 
MR-R & 10 & 5,436 & 3,232 & 2,576 \\ 
O-RS & 28 & 12,627 & 6,592 & 5,250 \\ 
WS-P & 30 & 14,099 & 7,728 & 6,151 \\ 
Y-GM & 9 & 952 & 188 & 141 \\ 
YM-Y & 17 & 8,794 & 6,269 & 4,997 \\ 
\hline \textbf{Total} & 128 & 64,791 & 35,825 & 28,536 \\  
\hline
\end{tabular}
\label{tab:ds1_detail}
\end{center}
\end{table}

\begin{table}[htbp]
\caption{Results for Dataset B}
\begin{center}
\begin{tabular}{lrrrr}
\hline
\textbf{Band} & \textbf{{Videos}}& 
\textbf{{Total Frames}}&\textbf{{K\=ak\=a} Frames} &
\textbf{Low Motion Frames} \\
\hline
-Y	&680	&74,780	&38,713	&30,264\\
K-WP	&67	&15,179	&10,283	&8,156\\
L-MB	&37	&4,506	&2,721	&2,135\\
L-XX	&96	&9,503	&5,466	&4,261\\
L-YM	&39	&3,011	&1,665	&1,284\\
LM-G	&135	&14,863	&8,447	&6,613\\
MO-R	&140	&27,840	&19,220	&15,231\\
MR-L	&5	&1,308	&957	&760\\
O-LW	&145	&15,208	&10,260	&8,049\\
P-XX	&11	&1,091	&773	&608\\
PR-K	&50	&4,796	&2,398	&1,866\\
W-BB	&13	&1,091	&462	&356\\
WX-P	&249	&47,233	&27,095	&21,423\\
Y-GL	&133	&10,228	&5,619	&4,354\\
YM-Y	&21	&4,274	&3,027	&2,399\\
\hline \textbf{Total}	&1,821	&234,911	&137,106	&107,759\\
\hline
\end{tabular}
\label{tab:ds2_detail}
\end{center}
\end{table}

\begin{table}[htbp]
\caption{Results for Dataset C}
\begin{center}
\begin{tabular}{lrrrr}
\hline
\textbf{Band} & \textbf{{Videos}}& 
\textbf{{Total Frames}}&\textbf{{K\=ak\=a} Frames} &
\textbf{Low Motion Frames} \\
\hline
-Y	&136	&125,706	&56,425	&45,001\\
K-PW	&8	&4,601	&1,950	&1,552\\
KR-L	&39	&55,304	&23,826	&19,019\\
L-MB	&24	&37,241	&17,410	&13,906\\
LM-G	&9	&12,125	&6,222	&4,970\\
O-LW	&10	&14,865	&8,038	&6,422\\
OL-Y	&2	&1,146	&319	&253\\
RM-X	&21	&20,570	&8,109	&6,470\\
W-GL	&71	&44,016	&20,751	&16,523\\
WX-P	&45	&72,683	&29,378	&23,455\\
X-XX	&31	&19,513	&9,153	&7,288\\
XX-O	&4	&4,823	&2,185	&1,744\\
Y-GL	&28	&29,591	&12,679	&10,113\\
YM-Y	&89	&62,063	&34,905	&27,835\\
	\hline \textbf{Total}&517	&504,247	&231,350	&184,551\\

\hline
\end{tabular}
\label{tab:ds3_detail}
\end{center}
\end{table}

\newpage
\section{Image Embedding Visualisations for Datasets B and C}
\label{appx:emb}

\begin{figure}[h!]
    \centering
    \begin{minipage}{0.45\textwidth}
        \centering
        \includegraphics[width=\linewidth]{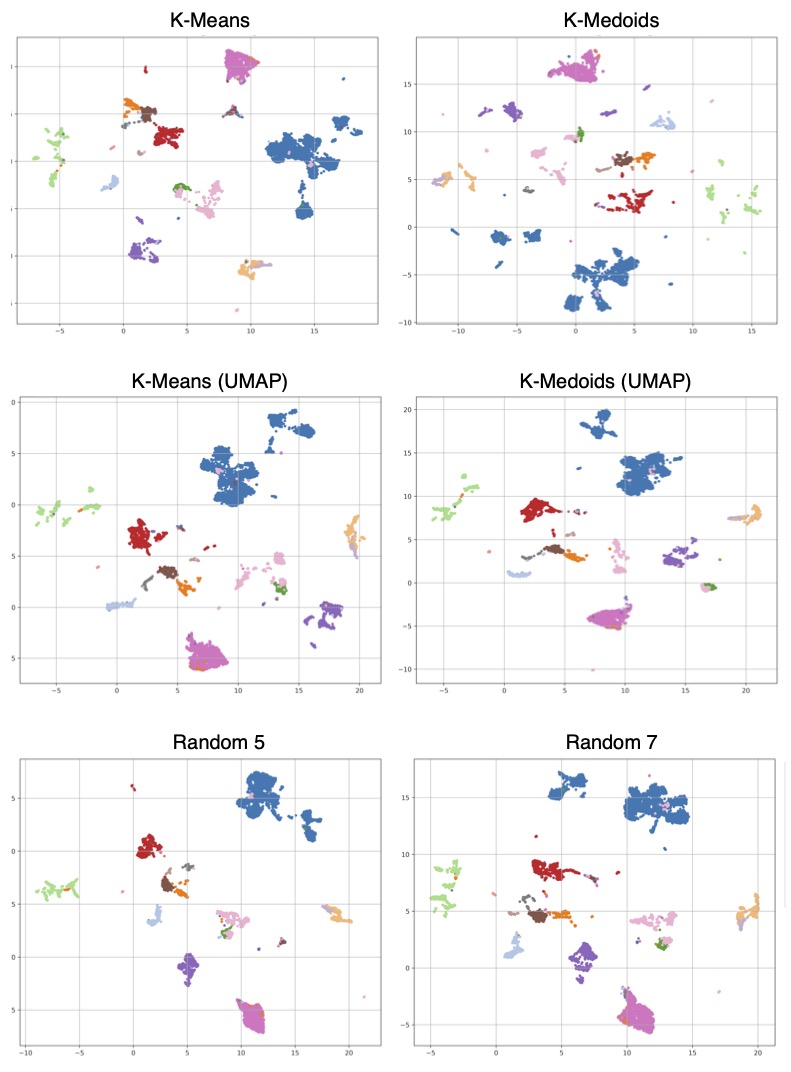}
        \caption{DINOv2 Image embeddings from Dataset B.}
        \label{fig:emb_ds2}
    \end{minipage}
    \hfill
    \begin{minipage}{0.45\textwidth}
        \centering
        \includegraphics[width=\linewidth]{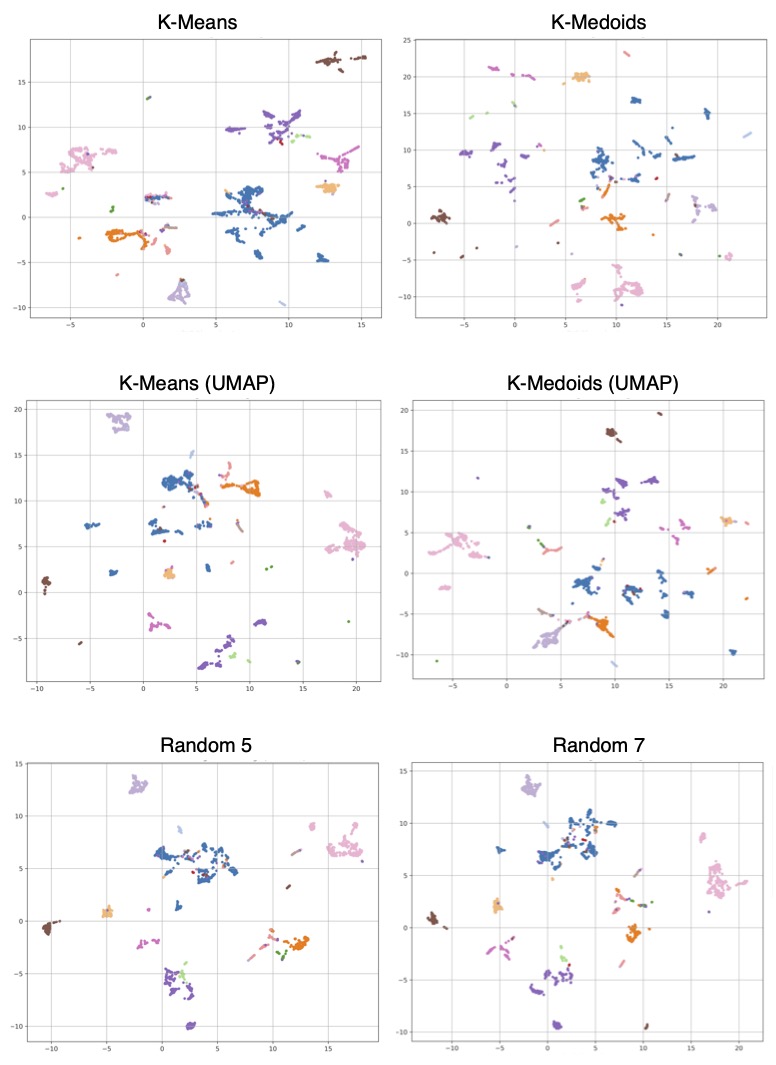}
        \caption{DINOv2 Image embeddings from Dataset C.}
        \label{fig:emb_ds3}
    \end{minipage}
\end{figure}

\ifCLASSOPTIONcaptionsoff
  \newpage
\fi

\newpage
\bibliographystyle{IEEEtran} 
\bibliography{refs}

\begin{thebibliography}{10}
\providecommand{\url}[1]{#1}
\csname url@samestyle\endcsname
\providecommand{\newblock}{\relax}
\providecommand{\bibinfo}[2]{#2}
\providecommand{\BIBentrySTDinterwordspacing}{\spaceskip=0pt\relax}
\providecommand{\BIBentryALTinterwordstretchfactor}{4}
\providecommand{\BIBentryALTinterwordspacing}{\spaceskip=\fontdimen2\font plus
\BIBentryALTinterwordstretchfactor\fontdimen3\font minus \fontdimen4\font\relax}
\providecommand{\BIBforeignlanguage}[2]{{%
\expandafter\ifx\csname l@#1\endcsname\relax
\typeout{** WARNING: IEEEtran.bst: No hyphenation pattern has been}%
\typeout{** loaded for the language `#1'. Using the pattern for}%
\typeout{** the default language instead.}%
\else
\language=\csname l@#1\endcsname
\fi
#2}}
\providecommand{\BIBdecl}{\relax}
\BIBdecl

\bibitem{Anthony2000}
L.~L. Anthony and D.~T. Blumstein, ``Integrating behaviour into wildlife conservation: the multiple ways that behaviour can reduce ne,'' \emph{Biological Conservation}, vol.~95, no.~3, pp. 303--315, 2000.

\bibitem{Nathan2008}
R.~Nathan, W.~M. Getz, E.~Revilla, M.~Holyoak, R.~Kadmon, D.~Saltz, and P.~E. Smouse, ``A movement ecology paradigm for unifying organismal movement research,'' \emph{Proceedings of the National Academy of Sciences of the United States of America}, vol. 105, pp. 19\,052--19\,059, 2008.

\bibitem{Stephenson2019}
P.~J. Stephenson, ``Integrating remote sensing into wildlife monitoring for conservation,'' \emph{Environmental Conservation}, vol.~46, pp. 181--183, 2019.

\bibitem{Debicki2021}
I.~T. D\k{e}bicki, E.~A. Mittell, B.~K. Kristjánsson, C.~A. Leblanc, M.~B. Morrissey, and K.~Terzić, ``Re-identification of individuals from images using spot constellations: a case study in arctic charr (salvelinus alpinus),'' \emph{Royal Society Open Science}, vol.~8, no.~7, p. 201768, 2021.

\bibitem{White2012}
T.~H. White, N.~J. Collar, R.~J. Moorhouse, V.~Sanz, E.~D. Stolen, and D.~J. Brightsmith, ``Psittacine reintroductions: Common denominators of success,'' \emph{Biological Conservation}, vol. 148, pp. 106--115, 2012.

\bibitem{Vidal2021}
M.~Vidal, N.~Wolf, B.~Rosenberg, B.~P. Harris, and A.~Mathis, ``Perspectives on individual animal identification from biology and computer vision,'' \emph{Integrative and Comparative Biology}, vol.~61, pp. 900--916, 2021.

\bibitem{Schneider2019}
S.~Schneider, G.~W. Taylor, S.~Linquist, and S.~C. Kremer, ``Past, present and future approaches using computer vision for animal re-identification from camera trap data,'' \emph{Methods in Ecology and Evolution}, vol.~10, pp. 461--470, 2019.

\bibitem{Li2020}
S.~Li, J.~Li, H.~Tang, R.~Qian, and W.~Lin, ``{ATRW: A benchmark for Amur tiger re-identification in the wild},'' in \emph{Proceedings of the 28th {ACM} International Conference on Multimedia}.\hskip 1em plus 0.5em minus 0.4em\relax ACM, 2020.

\bibitem{Crouse2017}
D.~Crouse, R.~L. Jacobs, Z.~Richardson, S.~Klum, A.~Jain, A.~L. Baden, and S.~R. Tecot, ``Lemurfaceid: a face recognition system to facilitate individual identification of lemurs,'' \emph{BMC Zoology}, vol.~2, no.~1, 2017.

\bibitem{Weinstein2018}
B.~G. Weinstein and C.~G.~B. Weinstein, ``A computer vision for animal ecology,'' \emph{Journal of Animal Ecology}, vol.~87, pp. 533--545, 2018.

\bibitem{Lurig2021}
M.~D. Lürig, S.~Donoughe, E.~I. Svensson, A.~Porto, and M.~Tsuboi, ``Computer vision, machine learning, and the promise of phenomics in ecology and evolutionary biology,'' \emph{Frontiers in Ecology and Evolution}, vol.~9, 2021.

\bibitem{Borowiec2022}
M.~L. Borowiec, R.~B. Dikow, P.~B. Frandsen, A.~McKeeken, G.~Valentini, and A.~E. White, ``Deep learning as a tool for ecology and evolution,'' \emph{Methods in Ecology and Evolution}, vol.~13, pp. 1640--1660, 2022.

\bibitem{Miele2021}
V.~Miele, G.~Dussert, B.~Spataro, S.~Chamaillé-Jammes, D.~Allainé, and C.~Bonenfant, ``Revisiting animal photo-identification using deep metric learning and network analysis,'' \emph{Methods in Ecology and Evolution}, vol.~12, pp. 863--873, 2021.

\bibitem{Tuia2022}
\BIBentryALTinterwordspacing
D.~Tuia, B.~Kellenberger, S.~Beery, B.~R. Costelloe, S.~Zuffi, B.~Risse, A.~Mathis, M.~W. Mathis, F.~van Langevelde, T.~Burghardt, R.~Kays, H.~Klinck, M.~Wikelski, I.~D. Couzin, G.~van Horn, M.~C. Crofoot, C.~V. Stewart, and T.~Berger-Wolf, ``Perspectives in machine learning for wildlife conservation,'' \emph{Nature Communications}, vol.~13, p. 792, 2022. [Online]. Available: \url{https://doi.org/10.1038/s41467-022-27980-y}
\BIBentrySTDinterwordspacing

\bibitem{openanimals2024}
S.~Hou, P.~Huang, Z.~Wang, Y.~Liu, Z.~Li, M.~Zhang, and Y.~Huang, ``Openanimals: Revisiting person re-identification for animals towards better generalization,'' \emph{arXiv preprint arXiv:2410.00204}, 2024.

\bibitem{Gu2022}
X.~Gu, H.~Chang, B.~Ma, and S.~Shan, ``Motion feature aggregation for video-based person re-identification,'' \emph{IEEE Transactions on Image Processing}, vol.~31, pp. 3908--3919, 2022.

\bibitem{Selvan2025}
C.~Selvan, H.~Anwar~Basha, K.~Meenakshi, and S.~Naveen, ``A review on person re-identification techniques and its analysis,'' \emph{IEEE Access}, vol.~13, pp. 22\,133--22\,145, 2025.

\bibitem{karanam2019}
S.~Karanam, M.~Gou, Z.~Wu, A.~Rates-Borras, O.~Camps, and R.~J. Radke, ``A systematic evaluation and benchmark for person re-identification: Features, metrics, and datasets,'' \emph{IEEE Transactions on Pattern Analysis and Machine Intelligence}, vol.~41, no.~3, pp. 523--536, 2019.

\bibitem{fintan2023}
F.~O'Sullivan, K.-R. Escott, R.~C. Shaw, and A.~Lensen, ``Feature-based image matching for identifying individual k\=ak\=a,'' \emph{arXiv preprint arXiv:2301.06678}, 2023.

\bibitem{Maddigan2024}
P.~Maddigan, O.~Ehrhardt, A.~Lensen, and R.~C. Shaw, ``Re-identification of individual k\={a}k\={a}: An explainable dino-based model,'' in \emph{2024 39th International Conference on Image and Vision Computing New Zealand (IVCNZ)}, 2024, pp. 1--6.

\bibitem{Xiao2023}
S.~Xiao, Y.~Wang, A.~Perkes, B.~Pfrommer, M.~Schmidt, K.~Daniilidis, and M.~Badger, ``Multi-view tracking, re-id, and social network analysis of a flock of visually similar birds in an outdoor aviary,'' \emph{International Journal of Computer Vision}, vol. 131, no.~6, pp. 1532--1549, 2023.

\bibitem{Ferreira2020}
A.~C. Ferreira, L.~R. Silva, F.~Renna, H.~B. Brandl, J.~P. Renoult, D.~R. Farine, R.~Covas, and C.~Doutrelant, ``Deep learning‐based methods for individual recognition in small birds,'' \emph{Methods in Ecology and Evolution}, vol.~11, no.~9, pp. 1072--1085, 2020.

\bibitem{Rogers2024}
M.~Rogers, K.~Knowles, G.~Gendron, S.~Heidari, D.~A.~S. Valdez, M.~Azhar, P.~O'Leary, S.~Eyre, M.~Witbrock, and P.~Delmas, ``Recurrence over video frames (rovf) for the re-identification of meerkats,'' \emph{arXiv preprint arXiv:2406.13002}, 2024.

\bibitem{Zuerl2023}
M.~Zuerl, R.~Dirauf, F.~Koeferl, N.~Steinlein, J.~Sueskind, D.~Zanca, I.~Brehm, L.~v. Fersen, and B.~Eskofier, ``Polarbearvidid: A video-based re-identification benchmark dataset for polar bears,'' \emph{Animals}, vol.~13, no.~5, 2023.

\bibitem{Schofield2023}
D.~Schofield, A.~Nagrani, A.~Zisserman, M.~Hayashi, T.~Matsuzawa, D.~Biro, and S.~Carvalho, ``Chimpanzee face recognition from videos in the wild using deep learning,'' \emph{Science Advances}, vol.~5, p. eaaw0736, 2023.

\bibitem{Wang2021}
L.~Wang, R.~Ding, Y.~Zhai, Q.~Zhang, W.~Tang, N.~Zheng, and G.~Hua, ``Giant panda identification,'' \emph{IEEE Transactions on Image Processing}, vol.~30, pp. 2837--2849, 2021.

\bibitem{Liu2022}
T.~Liu, Q.~Meng, J.-J. Huang, A.~Vlontzos, D.~Rueckert, and B.~Kainz, ``Video summarization through reinforcement learning with a 3d spatio-temporal u-net,'' \emph{IEEE Transactions on Image Processing}, vol.~31, pp. 1573--1586, 2022.

\bibitem{SavranKiziltepe2023}
R.~Savran~Kızıltepe, J.~Q. Gan, and J.~J. Escobar, ``A novel keyframe extraction method for video classification using deep neural networks,'' \emph{Neural Computing and Applications}, vol.~35, no.~34, pp. 24\,513--24\,524, 2023.

\bibitem{Kaur2024}
S.~Kaur, L.~Kaur, and M.~Lal, ``An effective key frame extraction technique based on feature fusion and fuzzy-c means clustering with artificial hummingbird,'' \emph{Scientific Reports}, vol.~14, no.~1, p. 26651, 2024.

\bibitem{Tang2023}
H.~Tang, L.~Ding, S.~Wu, B.~Ren, N.~Sebe, and P.~Rota, ``Deep unsupervised key frame extraction for efficient video classification,'' \emph{ACM Transactions on Multimedia Computing, Communications, and Applications}, vol.~19, no.~3, pp. 1--17, 2023.

\bibitem{Tan2024}
K.~Tan, Y.~Zhou, Q.~Xia, R.~Liu, and Y.~Chen, ``Large model based sequential keyframe extraction for video summarization,'' in \emph{Proceedings of the International Conference on Computing, Machine Learning and Data Science}, ser. CMLDS '24.\hskip 1em plus 0.5em minus 0.4em\relax New York, NY, USA: Association for Computing Machinery, 2024.

\bibitem{Sarwar2023}
M.~A. Sarwar, Y.-C. Lin, Y.-A. Daraghmi, T.-U. İK, and Y.-L. Li, ``Skeleton based keyframe detection framework for sports action analysis: Badminton smash case,'' \emph{IEEE Access}, vol.~11, pp. 90\,891--90\,900, 2023.

\bibitem{Yan2022}
G.~Yan and M.~Woźniak, ``Accurate key frame extraction algorithm of video action for aerobics online teaching,'' \emph{Mobile Networks and Applications}, vol.~27, no.~3, pp. 1252--1261, 2022.

\bibitem{Mitra2021}
A.~Mitra, S.~P. Mohanty, P.~Corcoran, and E.~Kougianos, ``A machine learning based approach for deepfake detection in social media through key video frame extraction,'' \emph{SN Computer Science}, vol.~2, no.~2, p.~98, 2021.

\bibitem{Athira2022}
P.~Athira, C.~Sruthi, and A.~Lijiya, ``A signer independent sign language recognition with co-articulation elimination from live videos: An indian scenario,'' \emph{Journal of King Saud University - Computer and Information Sciences}, vol.~34, no.~3, pp. 771--781, 2022.

\bibitem{Lu2022}
Z.~Lu, G.~Zhang, G.~Huang, Z.~Yu, C.-M. Pun, W.~Zhang, J.~Chen, and W.-K. Ling, ``Video person re-identification using key frame screening with index and feature reorganization based on inter-frame relation,'' \emph{International Journal of Machine Learning and Cybernetics}, vol.~13, no.~9, pp. 2745--2761, 2022.

\bibitem{Xie2022}
P.~Xie, X.~Xu, Z.~Wang, and T.~Yamasaki, ``Sampling and re-weighting: Towards diverse frame aware unsupervised video person re-identification,'' \emph{IEEE Transactions on Multimedia}, vol.~24, pp. 4250--4261, 2022.

\bibitem{Tao2023}
H.~Tao, Q.~Duan, and J.~An, ``An adaptive interference removal framework for video person re-identification,'' \emph{IEEE Transactions on Circuits and Systems for Video Technology}, vol.~33, no.~9, pp. 5148--5159, 2023.

\bibitem{cermak2024}
V.~Cermak, L.~Picek, L.~Adam, L.~Neumann, and J.~Matas, ``Wildfusion: Individual animal identification with calibrated similarity fusion,'' \emph{arXiv preprint arXiv:2408.12934}, 2024.

\bibitem{Wahltinez2024}
O.~Wahltinez and S.~J. Wahltinez, ``An open-source general purpose machine learning framework for individual animal re-identification using few-shot learning,'' \emph{Methods in Ecology and Evolution}, vol.~15, no.~2, pp. 373--387, 2024.

\bibitem{Adam_2024}
L.~Adam, V.~\v{C}erm\'ak, K.~Papafitsoros, and L.~Picek, ``Seaturtleid2022: A long-span dataset for reliable sea turtle re-identification,'' in \emph{Proceedings of the IEEE/CVF Winter Conference on Applications of Computer Vision (WACV)}, 2024, pp. 7146--7156.

\bibitem{Nepovinnykh2025}
E.~Nepovinnykh, V.~Immonen, T.~Eerola, C.~V. Stewart, and H.~Kälviäinen, ``Re-identification of patterned animals by multi-image feature aggregation and geometric similarity,'' \emph{IET Computer Vision}, vol.~19, no.~1, p. e12337, 2025.

\bibitem{Matlala2025}
B.~Matlala, D.~van~der Haar, and H.~Vandapalli, ``A novel approach to lion re-identification using vision transformers,'' in \emph{Artificial Intelligence Research}, A.~Gerber, J.~Maritz, and A.~W. Pillay, Eds.\hskip 1em plus 0.5em minus 0.4em\relax Cham: Springer Nature Switzerland, 2025, pp. 270--281.

\bibitem{Cermak2023}
V.~Čermák, L.~Picek, L.~Adam, and K.~Papafitsoros, ``Wildlifedatasets: An open-source toolkit for animal re-identification,'' 2023.

\bibitem{Ghaffari2024}
S.~Ghaffari, D.~W. Capson, K.~F. Li, and L.~Sielecki, ``Badger identification using handcrafted image matching with learned convolutional filter,'' in \emph{2024 IEEE 19th Conference on Industrial Electronics and Applications (ICIEA)}, 2024, pp. 1--5.

\bibitem{Li2019}
J.~Li, J.~Wang, Q.~Tian, W.~Gao, and S.~Zhang, ``Global-local temporal representations for video person re-identification,'' in \emph{Proceedings of the IEEE/CVF International Conference on Computer Vision (ICCV)}, October 2019.

\bibitem{Redmon2016}
J.~Redmon, S.~Divvala, R.~Girshick, and A.~Farhadi, ``You only look once: Unified, real-time object detection,'' in \emph{2016 IEEE Conference on Computer Vision and Pattern Recognition (CVPR)}, 2016, pp. 779--788.

\bibitem{khanam2024al}
R.~Khanam and M.~Hussain, ``Yolov11: An overview of the key architectural enhancements,'' \emph{arXiv preprint arXiv:2410.17725}, 2024.

\bibitem{yolo11_ultralytics}
\BIBentryALTinterwordspacing
G.~Jocher and J.~Qiu, ``Ultralytics yolo11,'' 2024. [Online]. Available: \url{https://github.com/ultralytics/ultralytics}
\BIBentrySTDinterwordspacing

\bibitem{jegham2025}
N.~Jegham, C.~Y. Koh, M.~Abdelatti, and A.~Hendawi, ``Yolo evolution: A comprehensive benchmark and architectural review of yolov12, yolo11, and their previous versions,'' \emph{arXiv preprint arXiv:2411.00201}, 2025.

\bibitem{liu2024}
S.~Liu, Z.~Zeng, T.~Ren, F.~Li, H.~Zhang, J.~Yang, Q.~Jiang, C.~Li, J.~Yang, H.~Su, J.~Zhu, and L.~Zhang, ``Grounding dino: Marrying dino with grounded pre-training for open-set object detection,'' 2024.

\bibitem{Padilla2020}
R.~Padilla, S.~L. Netto, and E.~A.~B. da~Silva, ``A survey on performance metrics for object-detection algorithms,'' in \emph{2020 International Conference on Systems, Signals and Image Processing (IWSSIP)}, 2020, pp. 237--242.

\bibitem{Farnebaeck2003}
G.~Farneb{\"a}ck, ``Two-frame motion estimation based on polynomial expansion,'' in \emph{Image Analysis}, J.~Bigun and T.~Gustavsson, Eds.\hskip 1em plus 0.5em minus 0.4em\relax Berlin, Heidelberg: Springer Berlin Heidelberg, 2003, pp. 363--370.

\bibitem{DINOv22024}
M.~Oquab, T.~Darcet, T.~Moutakanni, H.~Vo, M.~Szafraniec, V.~Khalidov, P.~Fernandez, D.~Haziza, F.~Massa, A.~El-Nouby, M.~Assran, N.~Ballas, W.~Galuba, R.~Howes, P.-Y. Huang, S.-W. Li, I.~Misra, M.~Rabbat, V.~Sharma, G.~Synnaeve, H.~Xu, H.~Jegou, J.~Mairal, P.~Labatut, A.~Joulin, and P.~Bojanowski, ``Dinov2: Learning robust visual features without supervision,'' 2024.

\bibitem{macqueen1967}
J.~MacQueen, ``Some methods for classification and analysis of multivariate observations,'' in \emph{Proceedings of the Fifth Berkeley Symposium on Mathematical Statistics and Probability, Volume 1: Statistics}, vol.~5.\hskip 1em plus 0.5em minus 0.4em\relax University of California press, 1967, pp. 281--298.

\bibitem{Park2009}
H.-S. Park and C.-H. Jun, ``A simple and fast algorithm for k-medoids clustering,'' \emph{Expert Systems with Applications}, vol.~36, no. 2, Part 2, pp. 3336--3341, 2009.

\bibitem{Domingos2012}
P.~Domingos, ``A few useful things to know about machine learning,'' \emph{Communications of the ACM}, vol.~55, no.~10, pp. 78--87, 2012.

\bibitem{mcinnes2020}
L.~McInnes, J.~Healy, and J.~Melville, ``Umap: Uniform manifold approximation and projection for dimension reduction,'' \emph{arXiv preprint arXiv:1802.03426}, 2020.

\bibitem{Allaoui2020}
M.~Allaoui, M.~L. Kherfi, and A.~Cheriet, ``Considerably improving clustering algorithms using umap dimensionality reduction technique: A comparative study,'' in \emph{Image and Signal Processing}, A.~El~Moataz, D.~Mammass, A.~Mansouri, and F.~Nouboud, Eds.\hskip 1em plus 0.5em minus 0.4em\relax Cham: Springer International Publishing, 2020, pp. 317--325.

\bibitem{Healy2024}
J.~Healy and L.~McInnes, ``Uniform manifold approximation and projection,'' \emph{Nature Reviews Methods Primers}, vol.~4, no.~1, p.~82, 2024.

\bibitem{Cristian2024}
P.-M. Cristian, V.-J. Aarón, E.-H.~D. Armando, M.-L.~Y. Estrella, N.-R. Daniel, G.-V. David, M.~Edgar, S.-C.~J. Paul, and R.-A. Osbaldo, ``Diffusion on {PCA-UMAP} manifold: The impact of data structure preservation to denoise high-dimensional single-cell {RNA} sequencing data.'' \emph{Biology}, vol.~13, 2024.

\bibitem{Rousseeuw1987}
P.~J. Rousseeuw, ``Silhouettes: A graphical aid to the interpretation and validation of cluster analysis,'' \emph{Journal of Computational and Applied Mathematics}, vol.~20, pp. 53--65, 1987.

\bibitem{fini2024}
E.~Fini, M.~Shukor, X.~Li, P.~Dufter, M.~Klein, D.~Haldimann, S.~Aitharaju, V.~G.~T. da~Costa, L.~Béthune, Z.~Gan, A.~T. Toshev, M.~Eichner, M.~Nabi, Y.~Yang, J.~M. Susskind, and A.~El-Nouby, ``Multimodal autoregressive pre-training of large vision encoders,'' \emph{arXiv preprint arXiv:2411.14402}, 2024.

\bibitem{McNemar1947}
Q.~McNemar, ``Note on the sampling error of the difference between correlated proportions or percentages,'' \emph{Psychometrika}, vol.~12, no.~2, p. 153–157, 1947.

\bibitem{Fisher2011}
M.~J. Fisher, A.~P. Marshall, and M.~Mitchell, ``Testing differences in proportions,'' \emph{Australian Critical Care}, vol.~24, no.~2, pp. 133--138, 2011.

\bibitem{Campello2013}
R.~J. G.~B. Campello, D.~Moulavi, and J.~Sander, ``Density-based clustering based on hierarchical density estimates,'' in \emph{Advances in Knowledge Discovery and Data Mining}, J.~Pei, V.~S. Tseng, L.~Cao, H.~Motoda, and G.~Xu, Eds.\hskip 1em plus 0.5em minus 0.4em\relax Berlin, Heidelberg: Springer Berlin Heidelberg, 2013, pp. 160--172.

\end{thebibliography}

\end{document}